\documentclass[final,5p,times,twocolumn,authoryear]{elsarticle}

\usepackage{amsmath, amsthm, amssymb}

\newtheorem{theorem}{Theorem}[section] 
\newtheorem{lemma}[theorem]{Lemma}        
\newtheorem{corollary}[theorem]{Corollary}   
\newtheorem{proposition}[theorem]{Proposition}
\newtheorem{definition}[theorem]{Definition}
\newtheorem{hypothesis}[theorem]{Hypothesis}

\usepackage{orcidlink}
\usepackage{booktabs}
\usepackage{float}
\usepackage{subfiles}
\usepackage{graphicx}
\usepackage{multirow}
\usepackage{geometry}
\usepackage{placeins}
\usepackage{balance}
\usepackage{subcaption}
\usepackage{tabularx}
\usepackage{xurl}
\usepackage{hyperref}
\hypersetup{
  hidelinks,
  pdftitle={Attraction-Ranking Mismatch in Out-of-Distribution Detection: Theory and Evidence from Hamiltonian Feature Dynamics},
  pdfauthor={Zilin Wang and Lianghai Xiao}
}
\usepackage{longtable}
\usepackage{array}
\usepackage{tikz}
\usepackage{pgfplots}
\pgfplotsset{compat=1.18}
\usepgfplotslibrary{groupplots}
\usetikzlibrary{arrows.meta,positioning,calc,decorations.pathmorphing,fit,backgrounds}
\journal{Knowledge-Based Systems}

\newcommand{\method}{Centroid--MSP}

\begin{document}
\raggedbottom

\begin{frontmatter}



\title{Attraction--Ranking Mismatch in Out-of-Distribution Detection: Theory and Evidence from Hamiltonian Feature Dynamics}

\author[label1]{Zilin Wang}
\author[label1]{Lianghai Xiao \corref{cor1}}
\cortext[cor1]{Corresponding author. \ead{xiaolh@jnu.edu.cn}} 

\affiliation[label1]{
organization={School of Information Science and Technology},
addressline={Jinan University}, 
city={Guangzhou},
postcode={510632}, 
country={China}}

\begin{abstract}
Attracting features toward learned in-distribution (ID) regions is an intuitive
strategy for out-of-distribution (OOD) detection, but does such attraction
improve detection? We investigate this question using class-conditioned
Hamiltonian dynamics as a controlled testbed and identify an
\emph{attraction--ranking mismatch}: improving individual class-affinity scores
need not improve the relative ordering of ID and OOD samples. Our analysis
separates misalignment between the driving field and the detection score from
ranking failure that persists even when they are aligned. An exact AUROC
decomposition and a local boundary analysis identify relative score changes
at the ID--OOD ranking boundary as the source of first-order AUROC change
under regularity conditions.
Experiments demonstrate that a score-aligned rule can improve performance on
a held-out source validation split yet consistently degrade it after transfer,
even without a single winning-class change. A separate perturbation study
finds that near-boundary relative score changes track the direction of small
AUROC changes in the tested settings. Finally, centroid approximation and shared scales connect the
zero-step score to a static centroid family that includes the existing CTM
detector. Together,
these results explain why class attraction is insufficient to justify feature
refinement and motivate evaluating its effect on the ranking boundary.
\end{abstract}


\end{frontmatter}


  
\section{Introduction}
\label{sec:Introduction}

Post-hoc out-of-distribution (OOD) detection asks whether a test sample belongs
to the distribution represented by a fixed classifier
\citep{hendrycks2016baseline,DBLP:journals/corr/LiangLS17}. Existing detectors
extract evidence from confidence, logits, or feature geometry
\citep{NEURIPS2020_f5496252,Wang_2022_CVPR,pmlr-v162-sun22d}. A recurring
intuition is that an in-distribution (ID) feature should lie near, or move
toward, a learned class region. The unresolved question is whether this
pointwise geometric attraction improves the \emph{relative ordering} of ID and
OOD samples.

We turn that intuition into a minimal dynamical problem. Normalized training
features define class-conditioned wells on the unit sphere; a query starts
from rest; and the strongest class affinity visited along its trajectory is
used as ID evidence. Hamiltonian dynamics provide a transparent testbed because
the force, trajectory, and observed score are all explicit. We use this model
to examine whether attraction implies better detection.

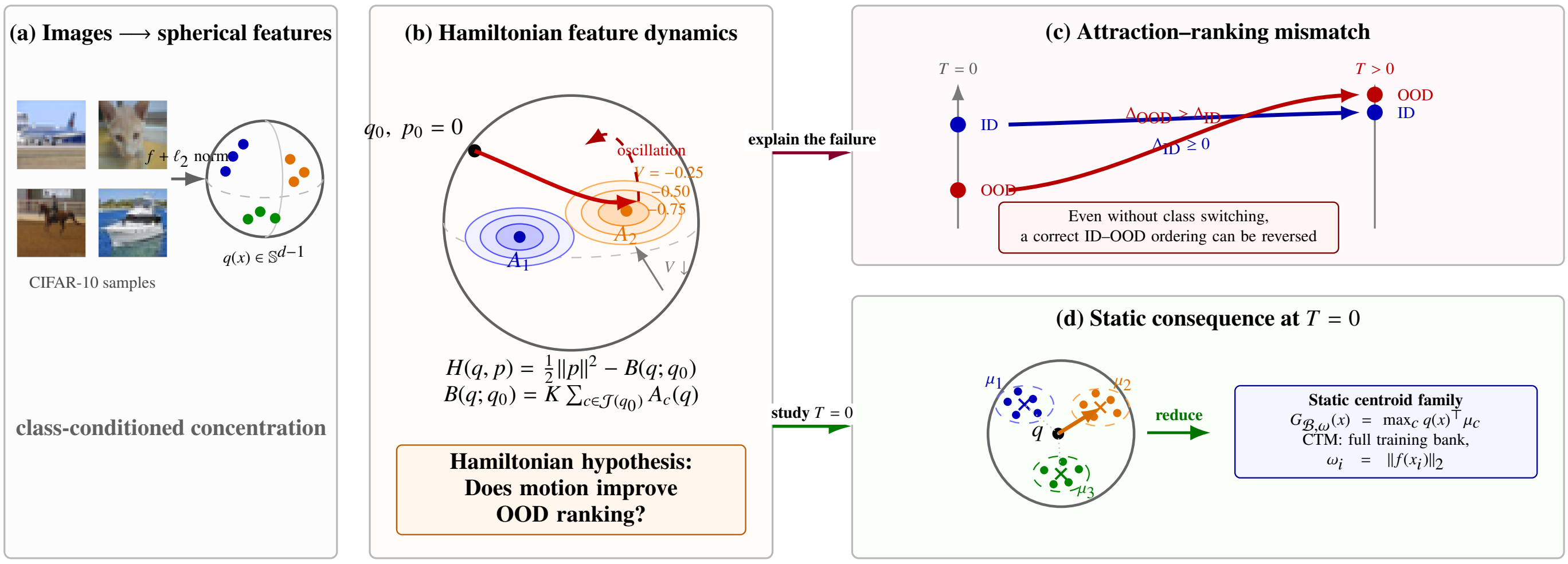
\begin{figure*}[t]
\centering
\resizebox{.985\textwidth}{!}{%
\begin{tikzpicture}[x=1cm,y=1cm,>=Latex,font=\scriptsize]
  \tikzset{
    panel/.style={draw=black!28,line width=.55pt,rounded corners=2pt,
      fill=white},
    title/.style={font=\scriptsize\bfseries,align=center},
    flow/.style={-{Latex[length=2.3mm,width=1.55mm]},line width=.85pt,
      draw=black!62},
    branch/.style={-{Latex[length=2.5mm,width=1.7mm]},line width=1pt},
    note/.style={rounded corners=1.6pt,align=center,inner sep=2.5pt}
  }

  \draw[panel,fill=black!1] (0,.15) rectangle (3.55,6.05);
  \draw[panel,fill=orange!2] (3.90,.15) rectangle (8.20,6.05);
  \draw[panel,fill=purple!2] (9.05,3.28) rectangle (16.65,6.05);
  \draw[panel,fill=green!2] (9.05,.15) rectangle (16.65,2.95);

  \node[title] at (1.775,5.74) {(a) Images $\longrightarrow$ spherical features};
  \node[inner sep=0pt,draw=black!40,line width=.35pt] at (.50,4.67)
    {\includegraphics[width=.72cm]{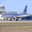}};
  \node[inner sep=0pt,draw=black!40,line width=.35pt] at (1.37,4.67)
    {\includegraphics[width=.72cm]{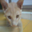}};
  \node[inner sep=0pt,draw=black!40,line width=.35pt] at (.50,3.73)
    {\includegraphics[width=.72cm]{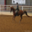}};
  \node[inner sep=0pt,draw=black!40,line width=.35pt] at (1.37,3.73)
    {\includegraphics[width=.72cm]{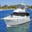}};
  \node[font=\tiny,align=center,text=black!72] at (.94,3.08)
    {CIFAR-10 samples};
  \draw[flow] (1.78,4.20)--node[above,font=\tiny]
    {$f$ + $\ell_2$ norm.} (2.16,4.20);

  \begin{scope}[shift={(2.78,4.20)}]
    \draw[thick,black!64] (0,0) circle (.62);
    \draw[black!28,dashed] (-.62,0) arc[start angle=180,end angle=360,
      x radius=.62,y radius=.20];
    \draw[black!22] (.00,-.62) arc[start angle=-90,end angle=90,
      x radius=.19,y radius=.62];
    \foreach \x/\y in {-.34/.24,-.23/.37,-.42/.08}
      \fill[blue!72!black] (\x,\y) circle (1.6pt);
    \foreach \x/\y in {.27/.21,.42/.07,.31/-.03}
      \fill[orange!88!black] (\x,\y) circle (1.6pt);
    \foreach \x/\y in {-.05/-.35,.11/-.42,-.18/-.43}
      \fill[green!55!black] (\x,\y) circle (1.6pt);
    \node[font=\tiny] at (0,-.82) {$q(x)\in\mathbb S^{d-1}$};
  \end{scope}
  \node[font=\scriptsize\bfseries,text=black!62] at (1.78,1.55)
    {class-conditioned concentration};

  \node[title] at (6.05,5.74) {(b) Hamiltonian feature dynamics};
  \begin{scope}[shift={(6.05,3.73)}]
    \draw[thick,black!62] (0,0) circle (1.36);
    \draw[black!25,dashed] (-1.36,0) arc[start angle=180,end angle=360,
      x radius=1.36,y radius=.37];
    \fill[blue!5] (-.55,-.15) ellipse (.58 and .31);
    \fill[blue!11] (-.55,-.15) ellipse (.42 and .23);
    \fill[blue!23] (-.55,-.15) ellipse (.25 and .14);
    \draw[blue!58] (-.55,-.15) ellipse (.58 and .31);
    \draw[blue!66] (-.55,-.15) ellipse (.42 and .23);
    \draw[blue!75] (-.55,-.15) ellipse (.25 and .14);
    \fill[orange!6] (.56,.11) ellipse (.62 and .34);
    \fill[orange!13] (.56,.11) ellipse (.45 and .24);
    \fill[orange!28] (.56,.11) ellipse (.27 and .14);
    \draw[orange!65] (.56,.11) ellipse (.62 and .34);
    \draw[orange!76] (.56,.11) ellipse (.45 and .24);
    \draw[orange!88] (.56,.11) ellipse (.27 and .14);
    \node[font=\tiny,text=orange!85!black] at (1.04,.54) {$V=-0.25$};
    \node[font=\tiny,text=orange!85!black] at (1.06,.34) {$-0.50$};
    \node[font=\tiny,text=orange!85!black] at (1.02,.15) {$-0.75$};
    \fill[blue!70!black] (-.55,-.15) circle (1.8pt) node[below=1pt] {$A_1$};
    \fill[orange!90!black] (.59,.13) circle (1.8pt) node[below=1pt] {$A_2$};
    \coordinate (qzero) at (-1.03,.77);
    \coordinate (qturn) at (.72,.23);
    \coordinate (qend) at (.14,.83);
    \fill[black] (qzero) circle (2pt) node[above left=0pt] {$q_0,\ p_0=0$};
    \draw[red!78!black,very thick,-{Latex[length=2.3mm]}]
      (qzero)..controls (-.45,.54) and (.14,.20)..(qturn);
    \draw[red!65!black,thick,dashed,-{Latex[length=2.2mm]}]
      (qturn)..controls (.73,.75) and (.45,1.01)..(qend);
    \node[font=\tiny,text=red!75!black] at (.86,.78) {oscillation};
    \draw[-{Latex[length=1.7mm]},black!52,line width=.55pt]
      (.98,-.72)--(.68,-.24) node[midway,right=1pt,font=\tiny] {$V\downarrow$};
  \end{scope}
  \node[font=\scriptsize,align=center] at (6.05,2.02)
    {$H(q,p)=\tfrac12\lVert p\rVert^2-B(q;q_0)$\\
     $B(q;q_0)=K\sum_{c\in\mathcal J(q_0)}A_c(q)$};
  \node[note,draw=orange!72!black,fill=orange!7,text width=3.55cm,
    font=\scriptsize\bfseries] at (6.05,.88)
    {Hamiltonian hypothesis:\\Does motion improve\\OOD ranking?};

  \draw[branch,draw=purple!68!black] (8.20,4.48)
    --node[above,font=\tiny\bfseries,fill=white,inner sep=1pt]{explain the failure}
    (9.05,4.48);
  \draw[branch,draw=green!48!black] (8.20,1.56)
    --node[above,font=\tiny\bfseries,fill=white,inner sep=1pt]{study $T=0$}
    (9.05,1.56);

  \node[title] at (12.85,5.75) {(c) Attraction--ranking mismatch};
  \node[font=\tiny\bfseries,text=black!65] at (10.18,5.38)
    {$T=0$};
  \draw[-{Latex[length=1.8mm]},black!52,line width=.55pt]
    (10.18,3.67)--(10.18,5.22);
  \fill[blue!72!black] (10.18,4.78) circle (2.4pt);
  \node[font=\tiny,text=blue!72!black,anchor=west] at (10.30,4.78)
    {ID};
  \fill[red!72!black] (10.18,4.08) circle (2.4pt);
  \node[font=\tiny,text=red!72!black,anchor=west] at (10.30,4.08)
    {OOD};

  \node[font=\tiny\bfseries,text=red!72!black] at (14.63,5.38)
    {$T>0$};
  \draw[-{Latex[length=1.8mm]},black!52,line width=.55pt]
    (14.63,3.67)--(14.63,5.22);
  \fill[blue!72!black] (14.63,4.91) circle (2.4pt);
  \node[font=\tiny,text=blue!72!black,anchor=west] at (14.75,4.91)
    {ID};
  \fill[red!72!black] (14.63,5.10) circle (2.4pt);
  \node[font=\tiny,text=red!72!black,anchor=west] at (14.75,5.10)
    {OOD};

  \draw[-{Latex[length=2.1mm]},very thick,blue!65!black]
    (10.72,4.78)..controls (12.05,4.82) and (13.05,4.87)..
    node[midway,below=2pt,font=\tiny,text=blue!65!black]
      {$\Delta_{\rm ID}\ge0$}
    (14.48,4.91);
  \draw[-{Latex[length=2.1mm]},very thick,red!72!black]
    (10.72,4.08)..controls (12.00,4.20) and (13.10,5.03)..
    node[pos=.48,above=2pt,font=\tiny,text=red!72!black]
      {$\Delta_{\rm OOD}>\Delta_{\rm ID}$}
    (14.48,5.10);
  \node[note,draw=red!48!black,fill=red!4,text width=3.45cm,font=\tiny]
    at (12.43,3.68)
    {Even without class switching,\\a correct ID--OOD ordering can be reversed};

  \node[title] at (12.85,2.69) {(d) Static consequence at $T=0$};
  \begin{scope}[shift={(11.28,1.48)}]
    \draw[thick,black!60] (0,0) circle (.78);
    \draw[blue!58,dashed] (-.39,.31) ellipse (.31 and .20);
    \foreach \x/\y in {-.57/.34,-.46/.45,-.30/.37,-.25/.22,-.48/.20}
      \fill[blue!72!black] (\x,\y) circle (1.25pt);
    \node[font=\scriptsize,text=blue!72!black] at (-.39,.31) {$\boldsymbol{\times}$};
    \node[font=\tiny,text=blue!72!black] at (-.70,.55) {$\mu_1$};

    \draw[orange!70,dashed] (.42,.28) ellipse (.31 and .20);
    \foreach \x/\y in {.24/.31,.37/.43,.53/.36,.60/.22,.35/.18}
      \fill[orange!88!black] (\x,\y) circle (1.25pt);
    \node[font=\scriptsize,text=orange!88!black] at (.42,.28) {$\boldsymbol{\times}$};
    \node[font=\tiny,text=orange!88!black] at (.67,.53) {$\mu_2$};

    \draw[green!52!black,dashed] (.00,-.43) ellipse (.30 and .19);
    \foreach \x/\y in {-.18/-.39,-.08/-.54,.08/-.52,.19/-.38,.02/-.30}
      \fill[green!52!black] (\x,\y) circle (1.25pt);
    \node[font=\scriptsize,text=green!48!black] at (.00,-.43) {$\boldsymbol{\times}$};
    \node[font=\tiny,text=green!48!black] at (.28,-.62) {$\mu_3$};

    \fill[black] (-.03,.00) circle (1.8pt) node[left=1pt] {$q$};
    \draw[black!25,densely dotted] (-.03,.00)--(-.39,.31);
    \draw[orange!84!black,very thick,-{Latex[length=2mm]}]
      (-.03,.00)--(.42,.28);
    \draw[black!25,densely dotted] (-.03,.00)--(.00,-.43);
  \end{scope}
  \draw[-{Latex[length=2.2mm]},line width=.85pt,green!45!black]
    (12.20,1.49)--node[above,font=\tiny\bfseries]{reduce} (12.88,1.49);
  \node[note,draw=blue!58!black,fill=blue!4,text width=3.05cm,font=\tiny]
    at (14.75,1.50)
    {\textbf{Static centroid family}\\
     $G_{\mathcal B,\omega}(x)=\max_c q(x)^\top\mu_c$\\
     CTM: full training bank,\\
     $\omega_i=\lVert f(x_i)\rVert_2$};

\end{tikzpicture}%
}
\caption{Research map. Class-conditioned concentration on the feature sphere
defines potential wells and a testable Hamiltonian hypothesis. Positive-time
attraction need not preserve ID--OOD ordering; the locked score-aligned audit
shows that the mismatch can persist without a winning-class switch. The exact
$T=0$ reference exposes a static geometry that reduces, after explicit
candidate expansion, centroid approximation, and scale sharing, to a centroid
family containing CTM as a full-bank member.}
\label{fig:overview_v2}
\end{figure*}

The implication fails in two logically distinct ways. First, the field that
moves a feature can be misaligned with the instantaneous class evidence used by
the detector: summing class forces may reduce the affinity of the currently
winning class. Second, even a score-aligned flow can raise OOD scores more than ID
scores near the baseline ranking boundary. We call these failures
\emph{attraction--ranking mismatch}. The distinction matters because AUROC is
a functional of ID--OOD pairs, not an average of samplewise score changes.

The central ranking result identifies the relative score change at the
ID--OOD ranking boundary that determines the first-order AUROC change under
regularity conditions. An exact decomposition into corrected and spoiled
pairs explains finite changes. A first-step analysis links the driving field
to the detection score, while an ordering-preservation result states the
limits of universal improvement by deterministic refinement.

Our experiments show that local score alignment does not ensure better OOD
ranking. A rule that improves AUROC on a held-out source split reduces it on
the target benchmarks when the refinement rule and step count are kept fixed.
This failure occurs without observed winning-class changes within the fixed
candidate sets, showing that class switching is not required to explain the
loss in ranking quality. In a separate perturbation study, near-boundary
relative score changes track the direction of small AUROC changes more
consistently than their global average. The experiments thus illustrate
ranking failure despite local alignment; they do not establish the
destructive class interference allowed by the theory.

The $T=0$ reference separates the static geometric information from the
effect of motion. After removing candidate truncation, we bound the error
of replacing each class field by a radial centroid score. Under shared
scales, this approximate score is exactly rank-equivalent to centroid cosine
similarity. CTM \citep{nguyen2023ctm} is an existing full-bank,
feature-norm-weighted member of the resulting family. This relation provides
a static baseline for studying the contribution of motion.

The contribution of this study is threefold:
\begin{itemize}
    \item first, we analyze the attraction–ranking mismatch by separating field–score 
    alignment from ID–OOD ranking and identifying the boundary condition that governs 
    first-order AUROC change under regularity conditions;
    \item second, we show through controlled experiments that local score alignment 
    does not guarantee better ranking, even without winning-class changes, and 
    examine the boundary-based explanation in a separate perturbation study;
    \item third, we connect zero-step anchor scores to a static centroid family 
    through approximation bounds and shared-scale ranking equivalence.
    This analysis also shows that CTM is an existing member of this family.
\end{itemize}

The rest of the paper is organized as follows. Section~\ref{sec:Related Work} positions the problem. Section~\ref{sec:fundamental_model}
states the fundamental model, Section~\ref{sec:mismatch_theory} develops the
theory, and Section~\ref{sec:Static Reduction and Detector} derives its static
consequence. Section~\ref{sec:Experiments} tests the theoretical chain, followed
by discussion and limitations.

\section{Related Work}
\label{sec:Related Work}

\subsection{Post-hoc OOD scores and feature shaping}

Post-hoc methods leave the classifier fixed. Output-based scores include
maximum softmax probability (MSP),
ODIN, Energy, MaxLogit, generalized entropy, and LogitGap
\citep{hendrycks2016baseline,DBLP:journals/corr/LiangLS17,
NEURIPS2020_f5496252,hendrycks2022scaling,Liu_2023_CVPR,liang2025logitgap}.
Feature shaping instead modifies the representation used for scoring: ReAct
clips activations, DICE retains high-contribution connections, ASH prunes and
rescales activations, Scale rescales features, and RankFeat removes a dominant
rank-one component
\citep{sun2021react,sun2022dice,djurisic2023extremely,ICLR2024_46d943bc,
song2022rankfeat}. Gradient Short-Circuit studies post-hoc feature
intervention, whereas FEVER-OOD uses dimensionality reduction and
singular-value regularization to mitigate free-energy vulnerabilities
\citep{Gu_2025_ICCV,IsaacMedina_2025_ICCV}. These distinct training-time and
inference-time approaches motivate our central question: when can a
representation change be expected to improve ranking?

\subsection{Geometric and prototype-based detection}

Geometric detectors use class statistics, neighborhoods, or residual
subspaces. Representative examples include Mahalanobis distance, RMDS, KNN,
ViM, and SHE
\citep{NEURIPS2018_abdeb6f5,DBLP:journals/corr/abs-2106-09022,
pmlr-v162-sun22d,Wang_2022_CVPR,zhang2023she}. NNGuide and SIRC combine
geometric evidence with classifier confidence
\citep{park2023nnguide,xia2024sirc}. Recent prototype methods refine the class
representative or enrich its comparison rule: SPROD suppresses spurious
features, POT uses prototype transport with virtual outliers, GROOD introduces
an artificial OOD prototype, and uncertainty-aware refinement couples
prototype geometry with energy
\citep{Zohrabi2025SPROD,ke2025pot,ElAraby2025GROOD,WeiLiu2026UAPR}.
CADRef retains class-aware relative information beyond a single centroid
\citep{Ling_2025_CVPR}.

The CTM method is the closest static construction to our zero-step family
\citep{nguyen2023ctm}. It averages unnormalized training features class-wise,
normalizes the resulting means, and uses maximum cosine similarity. 
Section~\ref{sec:Static Reduction and Detector} 
bounds the centroid approximation error and shows that the resulting radial
centroid score is rank-equivalent to cosine similarity under shared scales.
CTM is recovered as a full-bank special case.

\subsection{From geometric attraction to ranking}

Manifold and class-concentration arguments motivate localized feature geometry
\citep{cayton2008algorithms,fefferman2016testing}, while spectral and
information-geometric analyses relate learned representations to gradient or
predictive geometry \citep{seleznova2026gradpca,nguyen2026fisherrao}. These
works explain why geometric structure can be informative. They do not imply
that motion generated by this structure improves a distributional ranking.

Hamiltonian mechanics supplies a clean counterexample framework: a scalar
potential generates coupled position--momentum motion
\citep{arnold1989mathematical}, whereas an OOD score evaluates a separate
observable. We analyze why attraction need not improve ranking by separating
(i) alignment between the dynamical driver and the observed score from
(ii) alignment between samplewise score changes and ID--OOD pair ordering.
The resulting pairwise decomposition, boundary derivative, and limitation on
universal improvement apply to score refinement beyond this Hamiltonian
implementation when their stated assumptions hold.

\section{A Hamiltonian Model of Feature Refinement}
\label{sec:fundamental_model}
\label{sec:Hamiltonian Hypothesis}

To make the distinction between attraction and ranking explicit, we
construct a model in which both the feature dynamics and the detection
score are defined by class affinities. Normalized ID training features
serve as anchors, and their affinities define a potential that drives
query features toward the learned class regions. We then specify the
Hamiltonian dynamics and the score evaluated along each trajectory,
providing a concrete setting for analyzing how feature motion affects
ID--OOD ranking. Further geometric motivation and details of the numerical
update schemes are provided in \ref{app:fundamental_model_details}.

\subsection{A class-induced landscape}
\label{sec:hamiltonian_motivation}

Let $f:\mathcal X\to\mathbb R^d$ be the penultimate representation of a
fixed classifier, with $f(x)\ne0$ on the populations under consideration, and
retain only feature direction,
\begin{equation}
 q(x)=\frac{f(x)}{\lVert f(x)\rVert_2}\in\mathbb S^{d-1}.
 \label{eq:normalized_feature}
\end{equation}
Let $C$ be the number of classes and $K\ge1$ the number of anchors per class.
For $c\in\{1,\ldots,C\}$ and $j\in\{1,\ldots,K\}$, let
$a_{cj}\in\mathbb S^{d-1}$ be normalized training features used as anchors,
$w_{cj}\ge0$ their weights, and $\sigma_{cj}>0$ bandwidths. A decreasing
radial kernel $\phi:[0,\infty)\to\mathbb R$ defines the class affinity
\begin{equation}
 A_c(q)=\frac{1}{K}\sum_{j=1}^{K}w_{cj}
 \phi\!\left(\frac{\lVert q-a_{cj}\rVert_2}{\sigma_{cj}}\right).
 \label{eq:class_affinity}
\end{equation}
The full-class reference potential is $V(q)=-\sum_{c=1}^{C}A_c(q)$. In computation, a
nonempty candidate set $\mathcal J(q_0)\subseteq\{1,\ldots,C\}$ is selected
at the initial feature $q_0$ and held fixed. The scalar driving the motion is
\begin{equation}
 B(q;q_0)=K\sum_{c\in\mathcal J(q_0)}A_c(q).
 \label{eq:candidate_field}
\end{equation}
The dynamics then use the conditional potential $-B(q;q_0)$.

\subsection{The Hamiltonian question}
\label{sec:hamiltonian_flow}

\begin{figure*}[t]
\centering
\includegraphics[width=.97\textwidth]{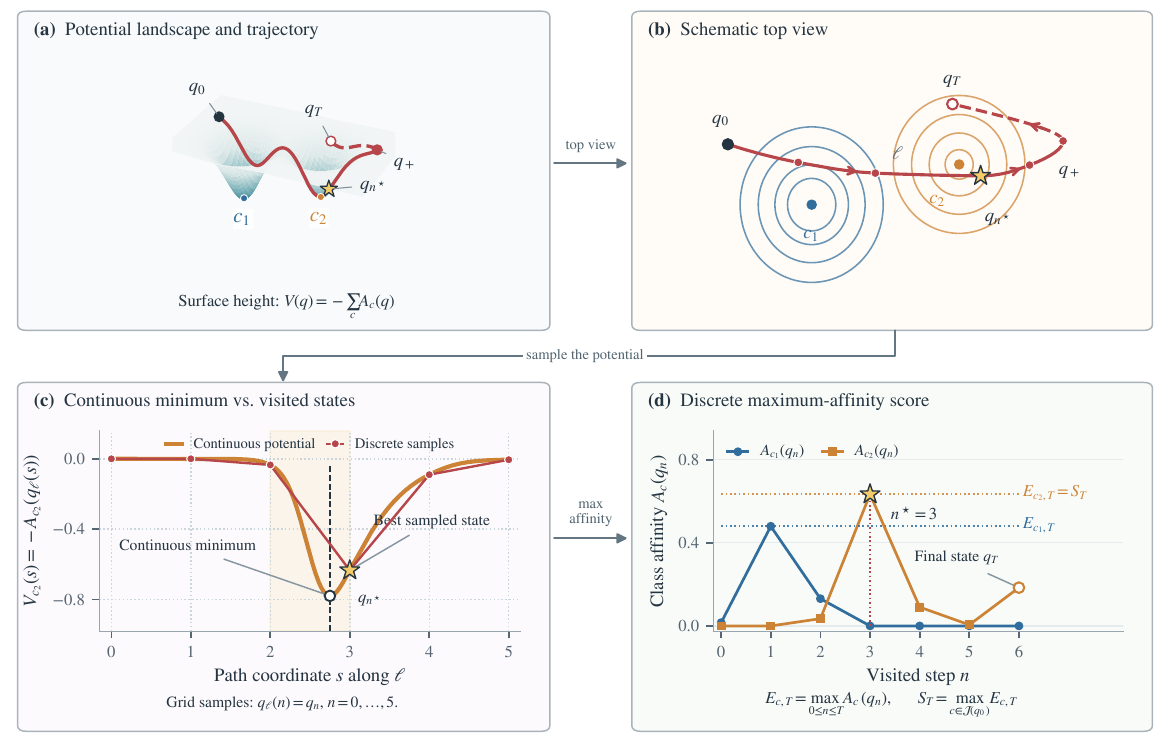}
\caption{Schematic construction of the trajectory score. (a) The surface
 height is $V(q)=-\sum_c A_c(q)$, with wells labeled $c_1,c_2$. (b) The same
 field and path in top view. Both views identify the initial state $q_0$,
 best sampled state $q_{n^\star}$, illustrative turning state $q_+$, and final
 state $q_T$; solid and dashed red curves show outward and returning motion.
 (c) Class potential along the outward branch $\ell$, parameterized by
 $q_\ell(n)=q_n$ for $n=0,\ldots,5$. Grid lines locate samples, and the red
 polygon joins their potential values. The continuous minimum (open circle)
 lies between samples and is lower than the best sampled value (star).
 (d) Each class contributes its maximum visited affinity $E_{c,T}$; their
 largest value defines $S_T$, which need not equal the final-state affinity.
 Stars mark the same state in all panels. All panels share affinities and
 sampled states; the planar motion is schematic, not a Hamiltonian simulation.}
\label{fig:potential_landscape_motivation_v5}
\end{figure*}

Let $P_q=I-qq^\top$. On the tangent bundle
$T\mathbb S^{d-1}=\{(q,p):\lVert q\rVert_2=1,\ q^\top p=0\}$, augment $q$
with momentum $p$ and start the query from rest under
\begin{equation}
\begin{aligned}
 H(q,p)&=\tfrac12\lVert p\rVert_2^2-B(q;q_0),\\
 \dot q&=p,\\
 \dot p&=P_q\nabla_qB(q;q_0)-\lVert p\rVert_2^2q,
 \qquad (q_0,p_0)=(q(x),0).
\end{aligned}
 \label{eq:hamilton_equations}
\end{equation}
The normal term preserves tangency. Let $\Delta t>0$ be the numerical step
size, $T\in\{0,1,\ldots\}$ the number of steps, and $q_n$ the feature after
step $n$. Define the instantaneous score and the maximum score along the
trajectory by
\begin{equation}
 m_{\mathcal J}(q;q_0)=\max_{c\in\mathcal J(q_0)}A_c(q),
 \qquad
 S_T(q_0)=\max_{0\le n\le T}m_{\mathcal J}(q_n;q_0).
 \label{eq:trajectory_score}
\end{equation}
Thus $S_T\ge S_0$ by construction even if instantaneous evidence decreases.
The implementation stores $K S_T$ because it sums rather than averages the
$K$ anchors; this common positive factor does not change rankings.
Figure~\ref{fig:potential_landscape_motivation_v5} shows the complete
construction: spherical class wells generate motion, while the detector
observes a maximum class affinity along the resulting trajectory.

\begin{hypothesis}[Hamiltonian attraction hypothesis]
\label{hyp:hamiltonian_attraction}
Within a specified field, step-size, and trajectory-length grid, nonzero
attraction toward the learned class regions improves the ID--OOD ranking
induced by the zero-step score $S_0$, and this advantage persists across
benchmark scale.
\end{hypothesis}

By construction, the trajectory score satisfies $S_T(q_0)\ge S_0(q_0)$
for every query. This samplewise guarantee does not establish the
hypothesis, because OOD scores may increase more than ID scores.
Moreover, the aggregate field driving the motion differs from the
maximum class affinity used for scoring. These distinctions motivate
the analysis in the next section.

\section{Attraction--Ranking Mismatch Theory}
\label{sec:mismatch_theory}

We examine two links between feature motion and detection performance.
First, does motion driven by the aggregate field increase the
instantaneous detection score? Second, do the resulting score changes
improve ID--OOD ranking? We begin with a step from rest to characterize
the local effect of the driving field, then turn to the relative
ordering of ID and OOD scores. Proofs and the connection between a
single trajectory step and the continuous score family used in the
boundary analysis are provided in \ref{app:attraction_ranking_proofs}.

\subsection{Field--score mismatch}
\label{sec:attraction_ranking_mismatch}

Let $P_q=I-qq^\top$ be the tangent projector of the unit sphere. Both the
constrained geodesic kick--drift--kick scheme and the original
position-normalized update have the same first-step expansion from rest;
\ref{app:fundamental_model_details} treats both schemes. Gradients
are taken with respect to the feature argument; the initial candidate set
is held fixed. Smoothness near a point on the sphere refers to a smooth
extension to an ambient neighborhood.

\begin{theorem}[First-step field--score mismatch]
\label{thm:first_step_mismatch}
Fix $\mathcal J(q_0)$ and suppose $B$ and $A_c$ are twice continuously
differentiable near $q_0$. Write
$b_0=P_{q_0}\nabla B(q_0;q_0)$ and
$g_c=P_{q_0}\nabla A_c(q_0)$. Starting from $p_0=0$,
\begin{equation}
 q_1=q_0+\frac{\Delta t^2}{2}b_0
 +O(\Delta t^4).
 \label{eq:first_step_expansion}
\end{equation}
Consequently,
\begin{align}
 B(q_1;q_0)-B(q_0;q_0)
 &=\frac{\Delta t^2}{2}\lVert b_0\rVert_2^2
 +O(\Delta t^4),
 \label{eq:total_field_local_gain}\\
 A_c(q_1)-A_c(q_0)
 &=\frac{\Delta t^2}{2}\langle g_c,b_0\rangle
 +O(\Delta t^4).
 \label{eq:class_field_local_gain}
\end{align}
Hence the aggregate driver's leading change is nonnegative, and is positive
when $b_0\ne0$, whereas an individual class affinity can either increase or
decrease.
\end{theorem}

The theorem makes the first mismatch explicit: intuitive attraction follows
the sum $B$, but the detector observes the largest class affinity. Other
classes can therefore pull against the class that currently provides the
score.

\begin{definition}[Class-interference index]
\label{def:gradient_interference}
Let $c^\star$ be the unique maximizer of $A_c(q_0)$ over
$\mathcal J(q_0)$ and set $g_c=P_{q_0}\nabla A_c(q_0)$. For
$g_{c^\star}\ne0$, define
\begin{equation}
 I(q_0)=\frac{\sum_{c\in\mathcal J(q_0)\setminus\{c^\star\}}
 \langle g_{c^\star},g_c\rangle}
 {\lVert g_{c^\star}\rVert_2^2}.
 \label{eq:gradient_interference}
\end{equation}
\end{definition}

\begin{corollary}[Destructive class interference]
\label{cor:gradient_interference}
Under Theorem~\ref{thm:first_step_mismatch} and
Definition~\ref{def:gradient_interference},
\begin{equation}
 A_{c^\star}(q_1)-A_{c^\star}(q_0)
 =\frac{K\Delta t^2}{2}\lVert g_{c^\star}\rVert_2^2
 [1+I(q_0)]+O(\Delta t^4).
 \label{eq:interference_expansion}
\end{equation}
If $I(q_0)<-1$, then for every sufficiently small nonzero $\Delta t$, the
aggregate field increases while the instantaneous maximum class affinity
decreases.
\end{corollary}

For these sufficiently small steps, $S_1(q_0)=S_0(q_0)$: the maximum along
the trajectory retains the initial score. The instantaneous score decreases,
but the trajectory score does not.

\begin{definition}[Driver--score alignment]
\label{def:driver_score_alignment}
For a differentiable driver $F$ and instantaneous score $s$ on the sphere, the
driver is \emph{locally score-aligned} at $q$ when
\begin{equation}
 \mathcal A_{F,s}(q)=
 \langle P_q\nabla s(q),P_q\nabla F(q)\rangle\ge0,
 \label{eq:driver_score_alignment}
\end{equation}
and strictly aligned when the inequality is strict.
\end{definition}

\begin{figure*}[t]
\centering
\includegraphics[width=.985\textwidth]{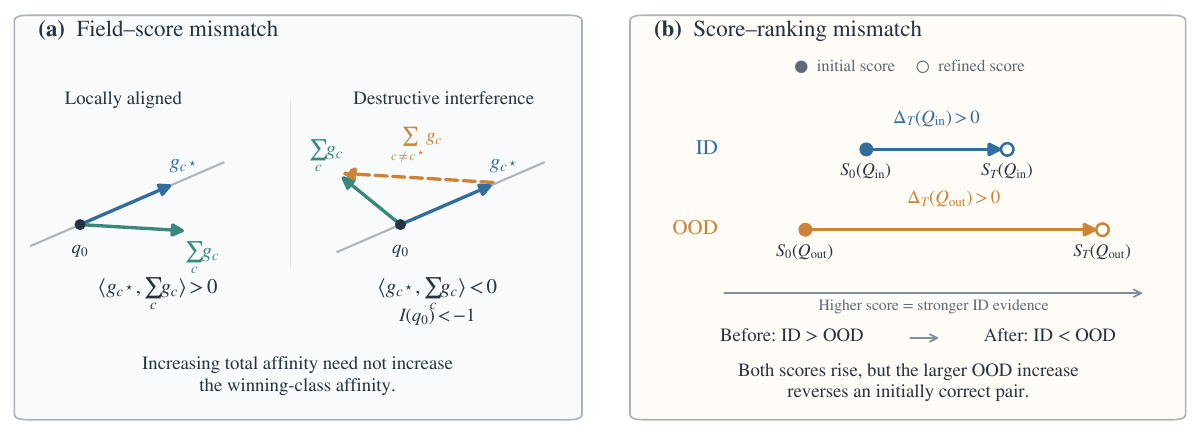}
\caption{Two levels of the attraction--ranking mismatch. (a) A projected
aggregate force can align with the winning-class score gradient (left), whereas
nonwinning classes can reverse that alignment and produce the destructive
condition $I(q_0)<-1$ (right). (b) Even when each sample's retained trajectory score is
nondecreasing, unequal ID and OOD increases can reverse an initially correct pair.
All class sums are over $\mathcal J(q_0)$. Panel (a) is a theoretical
schematic, not an observed audit sample; the geometry and values are
illustrative.}
\label{fig:field_score_mismatch}
\end{figure*}

For twice continuously differentiable $F$ and $s$, a step from rest changes
$s$ by
$\tfrac12\Delta t^2\mathcal A_{F,s}(q)+O(\Delta t^4)$. Choosing $F=s$
therefore removes the local field--score conflict away from ties. This is a
statement about a step from rest. At a general point on the continuous
trajectory,
\[
 \frac{d}{dt}s(q(t))=\langle P_{q(t)}\nabla s(q(t)),p(t)\rangle,
\]
so alignment of the driver alone does not imply a monotone instantaneous
score. The trajectory score $S_T$ is nondecreasing in $T$ because it retains
the maximum. Neither property determines the relative changes of ID and
OOD scores.
Figure~\ref{fig:field_score_mismatch} contrasts these two levels.

\subsection{From fixed winners to pairwise ranking}
\label{sec:trajectory_ranking}

For score-aligned dynamics, let
$c_0=\arg\max_{c\in\mathcal J(q_0)}A_c(q_0)$, using deterministic
tie-breaking. The initial-winner driver is
$F_{\rm init}=KA_{c_0}$, while the current-winner driver is
$F_{\rm cur}=K m_{\mathcal J}(\cdot;q_0)$ and reselects its active class at
each force evaluation. The initial-winner control retains only
$\max_{n\le T}A_{c_0}(q_n)$; the current-winner control uses $S_T$.

\begin{proposition}[Fixed-winner equivalence]
\label{prop:fixed_winner_equivalence}
Run the two systems from the same initial state with the same deterministic
kick--drift--kick scheme and candidate set. If $c_0$ is the unique maximizer at
every retained state of the initial-winner trajectory, then their states,
active classes, driver values, and retained scores are identical at every
step. For an integrator with additional internal force evaluations, the same
condition is required at those evaluation points.
\end{proposition}

When the winner stays fixed, these controls cannot differ through class
selection. Their common score can still change ID--OOD ordering. Define the
score increase
$\Delta_T(q)=S_T(q)-S_0(q)\ge0$.

Let $X_{\rm in}\sim P_{\rm in}$ and $X_{\rm out}\sim P_{\rm out}$ be
independent input samples, and set $Q_{\rm in}=q(X_{\rm in})$ and
$Q_{\rm out}=q(X_{\rm out})$. For a measurable input score
$s:\mathcal X\to\mathbb R$, larger values indicate stronger ID evidence. We use
\begin{equation}
\begin{aligned}
 \operatorname{AUROC}(s)
 &=\Pr\{s(X_{\rm in})>s(X_{\rm out})\}\\
 &\quad+\tfrac12\Pr\{s(X_{\rm in})=s(X_{\rm out})\}.
\end{aligned}
\label{eq:auroc_convention}
\end{equation}
For feature scores, write
$\operatorname{AUROC}_T=\operatorname{AUROC}(S_T\circ q)$.

\begin{definition}[Corrected and spoiled pairs]
\label{def:corrected_spoiled_pairs}
For the independent features defined above, let
\[
 M_0=S_0(Q_{\rm in})-S_0(Q_{\rm out}),\qquad
 D_T=\Delta_T(Q_{\rm in})-\Delta_T(Q_{\rm out}).
\]
The sets
\begin{equation}
\begin{aligned}
 \mathcal C_T&=\{M_0\le0,\ M_0+D_T>0\},\\
 \mathcal S_T&=\{M_0>0,\ M_0+D_T\le0\}.
\end{aligned}
 \label{eq:corrected_spoiled_sets}
\end{equation}
contain baseline errors corrected by the trajectory and baseline-correct pairs
spoiled by it, respectively.
\end{definition}

\begin{theorem}[Exact AUROC decomposition]
\label{thm:auroc_lift_decomposition}
Assume score ties have probability zero for $S_0$ and $S_T$, and write
$\Delta\operatorname{AUROC}=\operatorname{AUROC}_T-\operatorname{AUROC}_0$.
Then
\begin{equation}
\begin{aligned}
 \operatorname{AUROC}_T-\operatorname{AUROC}_0
 &=\Pr(\mathcal C_T)-\Pr(\mathcal S_T),\\
 |\Delta\operatorname{AUROC}|
 &\le\Pr(|M_0|\le|D_T|).
\end{aligned}
 \label{eq:auroc_pair_decomposition}
\end{equation}
\end{theorem}

The identity shows why nonnegative score increases are insufficient: ranking
improves only when corrected-pair mass exceeds spoiled-pair mass.

This finite-change identity applies to the trajectory score for any fixed
$T$. To identify which pairs control a small change, we next consider a
score family indexed by a continuous parameter $\eta$. Its expansion and
distributional assumptions must be checked separately; integer step counts
and maxima along a trajectory do not automatically satisfy them.

\begin{theorem}[Boundary-conditioned AUROC derivative]
\label{thm:boundary_auroc_derivative}
For the independent inputs $X_{\rm in},X_{\rm out}$ defined above, let
$s_\eta:\mathcal X\to\mathbb R$ be a measurable score family with
\[
\begin{aligned}
 s_\eta(x)&=s_0(x)+\eta r(x)+R_\eta(x),\\
 M_0&=s_0(X_{\rm in})-s_0(X_{\rm out}),\\
 U&=r(X_{\rm in})-r(X_{\rm out}).
\end{aligned}
\]
Suppose:
\begin{enumerate}
 \item $\varepsilon_\eta=\lVert R_\eta\rVert_{L^\infty(P_{\rm in}+P_{\rm out})}
 =o(|\eta|)$;
 \item $\mathbb E|U|<\infty$, and $(M_0,U)$ has a joint density
 $p_{M_0,U}(m,u)$ continuous in $m$ at zero for almost every $u$;
 \item for some $\delta>0$ and measurable $g\ge0$,
 $p_{M_0,U}(m,u)\le g(u)$ whenever $|m|\le\delta$, with
 $\int(1+|u|)g(u)\,du<\infty$, and
 $p_{M_0}(0)=\int p_{M_0,U}(0,u)\,du>0$.
\end{enumerate}
Then
\begin{equation}
 \left.\frac{d}{d\eta}\operatorname{AUROC}(s_\eta)\right|_{\eta=0}
 =p_{M_0}(0)\,\mathbb E[U\mid M_0=0].
 \label{eq:boundary_auroc_derivative}
\end{equation}
\end{theorem}

Here the conditional mean is defined by the density:
\[
 \mathbb E[U\mid M_0=0]
 =\frac{\int u\,p_{M_0,U}(0,u)\,du}{p_{M_0}(0)}.
\]
If the score family is defined only for $\eta\ge0$, the derivative is
right-sided. The ranking boundary $M_0=0$ consists of ID--OOD pairs with equal
baseline scores; it is not a classifier's class decision boundary. Relative
score change at this boundary controls the first-order AUROC change, whereas
the global mean $\mathbb E[U]$ need not have the same sign.

\begin{theorem}[No universal AUROC gain from deterministic refinement]
\label{thm:no_universal_auc_gain}
Let $s,s_{\rm ref}:\mathcal X\to\mathbb R$ be measurable baseline and
deterministic refined scores. If $s(x)>s(y)$ but
$s_{\rm ref}(x)\le s_{\rm ref}(y)$ for some $x,y$, then there exist ID
and OOD distributions for which $\operatorname{AUROC}(s_{\rm ref})<
\operatorname{AUROC}(s)$. Universal non-decrease over all ID/OOD distributions
without baseline ties therefore requires strict preservation of every ordering
induced by $s$; under that condition, the two AUROCs are equal for all such
distributions.
\end{theorem}

\begin{proof}
Choose the point-mass distributions $P_{\rm in}=\delta_x$ and
$P_{\rm out}=\delta_y$. Equation~\eqref{eq:auroc_convention} gives
$\operatorname{AUROC}(s)=1$, while
\[
 \operatorname{AUROC}(s_{\rm ref})=
 \begin{cases}
  0,&s_{\rm ref}(x)<s_{\rm ref}(y),\\
  1/2,&s_{\rm ref}(x)=s_{\rm ref}(y).
 \end{cases}
\]
Thus universal non-decrease requires
$s(x)>s(y)\Rightarrow s_{\rm ref}(x)>s_{\rm ref}(y)$ for every pair.
Applying this condition also with $x,y$ exchanged preserves the sign of every
nonzero baseline difference. If baseline ties have probability zero, then
\[
 \mathbf1_{\{s(X_{\rm in})>s(X_{\rm out})\}}
 =\mathbf1_{\{s_{\rm ref}(X_{\rm in})>s_{\rm ref}(X_{\rm out})\}}
 \quad\text{almost surely},
\]
and refined ties also have probability zero. Taking expectations proves
equality of the two AUROCs.
\end{proof}

The theorem limits guarantees over all ID/OOD distributions; it does not
rule out improvement on a particular task. Preserving every strict
baseline ordering leaves AUROC unchanged when baseline ties have
probability zero, whereas changing such an ordering can reduce AUROC
for some distributions.

\section{Static Scores at Zero Steps}
\label{sec:Static Reduction and Detector}

The preceding analysis studies how feature motion changes the ranking
relative to the initial score $S_0$. We now examine the structure of
this baseline: what geometric information does the detector use before
any motion takes place? Setting $T=0$ in
Eq.~\eqref{eq:trajectory_score} gives
$S_0(q)=\max_{c\in\mathcal J(q)}A_c(q)$.

To study its relation to centroid scoring, we consider the full-class
score $S_0^{\rm all}(q)=\max_{1\le c\le C}A_c(q)$.
Removing candidate truncation is an additional modeling choice and
need not preserve the original score. We first bound the error of
replacing each class field by a centroid score, then examine when
the resulting scores induce the same ranking as cosine similarity.
Proofs are provided in \ref{app:static_reduction_proofs}.

Write
$W_c=K^{-1}\sum_jw_{cj}$,
$\pi_{cj}=w_{cj}/\sum_\ell w_{c\ell}$,
$\bar a_c=\sum_j\pi_{cj}a_{cj}$, and
$\mu_c=\bar a_c/\lVert\bar a_c\rVert_2$. Assume
$\sum_jw_{cj}>0$ and $\bar a_c\ne0$ for every class considered.

\begin{proposition}[Centroid approximation]
\label{prop:centroid_approximation}
Suppose class $c$ uses a common bandwidth $\sigma_c$ and $\phi$ is
$L_\phi$-Lipschitz. Define
$\widetilde A_c(q)=W_c\phi(\lVert q-\mu_c\rVert_2/\sigma_c)$. Then
\begin{equation}
 |A_c(q)-\widetilde A_c(q)|\le
 \frac{W_cL_\phi}{\sigma_c}
 \left(\sum_j\pi_{cj}\lVert a_{cj}-\bar a_c\rVert_2
 +|1-\lVert\bar a_c\rVert_2|\right).
 \label{eq:spherical_centroid_bound}
\end{equation}
Consequently, the maximum-field error is bounded by the largest class-wise
right-hand side.
\end{proposition}

Let $\epsilon$ be the maximum, over classes, of the right-hand side of
Eq.~\eqref{eq:spherical_centroid_bound}, and write
$\widetilde S_0(q)=\max_c\widetilde A_c(q)$. Then
$\sup_q|S_0^{\rm all}(q)-\widetilde S_0(q)|\le\epsilon$. In particular,
\begin{equation}
\begin{aligned}
 S_0^{\rm all}(q)-S_0^{\rm all}(q')>2\epsilon
 \quad\Longrightarrow\quad
 \widetilde S_0(q)>\widetilde S_0(q').
\end{aligned}
\label{eq:centroid_margin_preservation}
\end{equation}
Smaller margins need not be preserved. The following exact equivalence
concerns the approximated centroid score, not the original anchor score.

\begin{proposition}[Common-scale ranking equivalence]
\label{prop:common_scale_equivalence}
If $W_c=W>0$ and $\sigma_c=\sigma>0$ are shared across classes and $\phi$ is
strictly decreasing, then for unit $q$ and $\mu_c$,
\begin{equation}
 \max_c\widetilde A_c(q)=W\phi\!\left(
 \frac{\sqrt{2-2G(q)}}{\sigma}\right),
 \qquad G(q)=\max_cq^\top\mu_c.
 \label{eq:radial_cosine_score_equivalence}
\end{equation}
The two scores select the same classes and rank all samples identically.
\end{proposition}

\begin{definition}[Static centroid family]
\label{def:static_centroid_family}
Let $\mathcal B=(\mathcal B_c)_{c=1}^C$, where $\mathcal B_c$ is a nonempty
set of training-example indices for class $c$. For positive directional weights
$\omega_i$ such that $\sum_{i\in\mathcal B_c}\omega_iq(x_i)\ne0$, define
\begin{equation}
\begin{aligned}
 \mu_c(\mathcal B,\omega)
 &=\frac{\sum_{i\in\mathcal B_c}\omega_iq(x_i)}
 {\lVert\sum_{i\in\mathcal B_c}\omega_iq(x_i)\rVert_2},\\
 G_{\mathcal B,\omega}(x)
 &=\max_c q(x)^\top\mu_c(\mathcal B,\omega).
\end{aligned}
 \label{eq:general_centroid_family}
\end{equation}
The collection of all such $G_{\mathcal B,\omega}$ is the static centroid
family induced by the zero-step reduction.
\end{definition}

Unequal bank sizes are also compatible with a common class multiplier.
For $K_c=|\mathcal B_c|$, replace $K^{-1}\sum_j$ in the class affinity by
$K_c^{-1}\sum_{i\in\mathcal B_c}$ and choose
\[
 w_{ci}=\frac{K_c W\omega_i}{\sum_{j\in\mathcal B_c}\omega_j}.
\]
Then $K_c^{-1}\sum_iw_{ci}=W$, while the normalized anchor weights are
$\pi_{ci}=\omega_i/\sum_j\omega_j$. Thus bank selection and directional
weighting identify different members of the same family under shared
bandwidths.

\begin{proposition}[CTM is a special case]
\label{prop:ctm_special_case}
Let $\mathcal B_{\rm all}$ contain all training indices for each class.
Write each training feature as $f(x_i)=r_iq(x_i)$ with
$r_i=\lVert f(x_i)\rVert_2$. CTM's normalized class mean is
$\mu_c(\mathcal B_{\rm all},r)$, and its maximum-cosine score is exactly
$G_{\mathcal B_{\rm all},r}$. Hence CTM is the full-bank,
feature-norm-weighted member of Definition~\ref{def:static_centroid_family}.
\end{proposition}

The reduction separates three operations: removing candidate truncation,
approximating each class field by a centroid, and imposing shared scales.
The approximation is controlled by an error bound; the resulting radial
centroid and cosine scores are exactly rank-equivalent under the shared-scale
conditions. CTM is an existing member of this static family.
This connection provides static controls for separating feature motion from
score construction; the ranking equivalence does not extend to the original
anchor field without accounting for the approximation error.

\section{Experiments}
\label{sec:Experiments}
\label{sec:Experimental Protocol}

The experiments address three questions: whether the tested attraction
improves OOD ranking, whether local score alignment prevents ranking failure,
and whether a boundary statistic explains the direction of small AUROC
changes. A final comparison separates the steps of the static reduction.
Destructive first-step interference is a theoretical possibility; the
real-data studies below do not establish its occurrence.

\subsection{Protocol}

We use OpenOOD v1.5 splits \citep{zhang2023openood}. CIFAR-10, CIFAR-100, and
ImageNet-200 use three official ResNet-18 checkpoints; ImageNet-1K uses a
pinned torchvision ResNet-50 V1, with DenseNet-121 as an additional backbone.
Tables~\ref{tab:per_dataset_cifar}--\ref{tab:per_dataset_imagenet} report the
individual Near-/Far-OOD datasets and results.

The original position-normalized dynamics use Gaussian anchor fields,
$\Delta t=0.05$, and $T\in\{0,1,3,10\}$. CIFAR uses 80 anchors per class and
all-class forces; ImageNet uses five anchors from 12 setup features per class
and a fixed initial top-20 candidate set. The separate score-aligned audit
uses five anchors from 12 setup features per class on every benchmark, all 10
classes on CIFAR-10, fixed top-20 candidates elsewhere, exact geodesic drift,
$\Delta t=0.05$, and $T\in\{0,1,3,5,10\}$. Current-winner, $T=5$ is selected on an
ImageNet-1K tuning split, checked on its disjoint holdout, and transferred
without retuning to CIFAR-10, CIFAR-100, and ImageNet-200. The score-aligned
rule and step count are fixed across benchmarks; each benchmark uses its own
classifier and anchor field. The score-aligned study and the separate
stability probe use distinct validation partitions, with parameter selection
and holdout evaluation kept separate. The probe step size is selected on
the source and fixed across targets. Neither study uses Near-/Far-OOD test loaders. The original trajectory comparisons and
static baselines report Near-/Far-OOD test results.
Proxy OOD is Tiny ImageNet for CIFAR and OpenImage-O for ImageNet.

The static comparisons use the same backbones and loaders. CTM uses the full
training bank, while the small-bank control uses 80 directions per class on
CIFAR and 12 on ImageNet. Its maximum cosine similarity is combined with MSP at
the previously selected weight $\alpha=0.8$ after ID-only standardization.
We call this the centroid--MSP baseline and use it to examine the static
construction in Section~\ref{sec:Static Reduction and Detector}.

We report AUROC and FPR@95; higher AUROC and lower FPR@95 are better. Following
the evaluation code, FPR@95 uses OOD as positive: it is the ID rejection rate
at $95\%$ OOD recall. AUROC assigns half credit to ties. CIFAR and ImageNet-200
results average three independently trained model seeds.
ImageNet-1K variability for anchor-based methods is over three sampled banks
on a fixed backbone. Selection and calibration settings are described in
\ref{app:implementation_details}.

\subsection{Testing the attraction hypothesis}
\label{sec:Testing Hamiltonian Dynamics}

Table~\ref{tab:trajectory_ablation} compares the zero-step Gaussian field with
the best nonzero step count in the tested grid. Small gains occur on CIFAR and
ImageNet-200, but every tested nonzero Gaussian trajectory is worse on
ImageNet-1K: even the retrospective best loses $1.63$ Near-OOD and $3.07$
Far-OOD AUROC points. The complete $T$ grid, IMQ control, five-kernel study,
and anchor-weighting ablation appear in
Tables~\ref{tab:trajectory_full}, \ref{tab:imagenet1k_t10_detail},
\ref{tab:potential_ablation}, and~\ref{tab:mass_full}. Nonzero attraction is
therefore not consistently better than $T=0$ across these benchmarks. This
comparison does not exclude improvements at other step sizes or trajectory lengths.

\begin{table}[!t]
\centering
\caption{Gaussian trajectory ablation. The best nonzero $T$ is selected
retrospectively from $\{1,3,10\}$. Entries are AUROC/FPR@95; $\Delta$ is the
AUROC change from $T=0$.}
\label{tab:trajectory_ablation}
\small
\setlength{\tabcolsep}{3.2pt}
\begin{tabular}{llcccc}
\toprule
ID & Group & $T=0$ & best $T>0$ & $T$ & $\Delta$ \\
\midrule
CIFAR-10 & Near & 88.59/42.36 & 88.95/40.91 & 10 & $+.36$ \\
         & Far  & 91.07/31.09 & 91.56/29.95 & 10 & $+.49$ \\
CIFAR-100& Near & 79.46/63.16 & 79.72/61.88 & 3 & $+.26$ \\
         & Far  & 83.41/51.10 & 84.01/49.61 & 3 & $+.60$ \\
IN-200   & Near & 79.44/64.42 & 80.22/61.89 & 3 & $+.78$ \\
         & Far  & 91.18/31.66 & 91.29/31.36 & 1 & $+.11$ \\
IN-1K    & Near & \textbf{66.14/77.84} & 64.51/79.27 & 1 & $-1.63$ \\
         & Far  & \textbf{77.44/65.16} & 74.37/69.75 & 3 & $-3.07$ \\
\bottomrule
\end{tabular}
\end{table}

\subsection{Ranking failure without class switching}
\label{sec:aligned_dynamics}

The current-winner flow raises AUROC from $71.79$ to $72.93$ on the
ImageNet-1K holdout, but this improvement does not transfer.
Table~\ref{tab:aligned_transfer} shows negative AUROC change on every seed
of the three target benchmarks. Spoiled pairs exceed
corrected pairs in all nine runs, as accounted for exactly by
Theorem~\ref{thm:auroc_lift_decomposition}. Moreover, none of the 20,205
audited sample--seed trajectories changes its initial winning class.
Thus ranking can deteriorate without a change of winning class.

Away from ties, the current-winner flow satisfies the local alignment
condition in Definition~\ref{def:driver_score_alignment}. Numerical checks
support the local expansion and confirm that the geodesic update preserves
the sphere and tangent momentum to numerical precision. The audit finds no
destructive first-step interference. \ref{app:aligned_dynamics_audit}
reports the numerical checks and the comparison with the initial-winner flow;
the exact equivalence in Proposition~\ref{prop:fixed_winner_equivalence}
additionally requires a unique winner at every force evaluation.

\begin{table*}[!t]
\centering
\caption{Validation-only score-alignment audit. ImageNet-1K is the holdout
after selection; other rows transfer the selected current-winner geodesic
$T=5$ rule without retuning. Entries are mean$\pm$standard deviation
AUROC/FPR@95; $\Delta_0$ is aligned $T=5$ AUROC minus $T=0$ AUROC.}
\label{tab:aligned_transfer}
\small
\setlength{\tabcolsep}{4.2pt}
\begin{tabular}{lcccc}
\toprule
Benchmark & $T=0$ & Aggregate $T=5$ & Current winner $T=5$ & $\Delta_0$ \\
\midrule
ImageNet-1K & $71.79\pm.82/72.63\pm3.17$ & $65.99\pm.23/74.59\pm1.40$ & $72.93\pm.57/72.03\pm2.66$ & $+1.14$ \\
CIFAR-10    & $89.42\pm.37/39.60\pm3.22$ & $90.13\pm.19/36.27\pm7.10$ & $87.98\pm.39/54.67\pm7.63$ & $-1.43$ \\
CIFAR-100   & $84.26\pm.88/47.07\pm6.88$ & $85.38\pm1.02/45.27\pm4.97$ & $83.84\pm.89/48.93\pm4.06$ & $-.42$ \\
ImageNet-200& $88.09\pm.69/38.78\pm2.67$ & $86.73\pm1.48/38.61\pm3.15$ & $87.53\pm.93/40.17\pm4.09$ & $-.56$ \\
\bottomrule
\end{tabular}
\end{table*}

\subsection{Evaluating the boundary-based diagnostic}
\label{sec:stability_probe_transfer}

To examine the local diagnostic suggested by
Theorem~\ref{thm:boundary_auroc_derivative}, we use a separate perturbation
with an explicit score derivative. A small repulsive step from a fixed class
centroid provides this control: its score decrease can be calculated for
each sample and compared across ID--OOD pairs. The probe uses the score at
the perturbed feature, not the maximum along an attractive trajectory. Let
$m(x)$ be MSP,
$\hat c(x)=\arg\max_{1\le c\le C}q(x)^\top\mu_c$,
$G(x)=q(x)^\top\mu_{\hat c(x)}$, with a deterministic rule for ties, and
$D(x)=(1-m(x))(1-G(x)^2)$. Freeze $m$, $\hat c$, and the ID-fit calibration
statistics, and set
\begin{equation}
\begin{aligned}
 q_\eta(x)&=\operatorname{Exp}_{q(x)}
 \!\left[-\eta(1-m(x))P_{q(x)}\mu_{\hat c(x)}\right],\\
 s_\eta^{\rm probe}(x)&=(1-\alpha)\frac{m(x)-\bar m}{\sigma_m}
 +\alpha\frac{q_\eta(x)^\top\mu_{\hat c(x)}-\bar G}{\sigma_G},
 \qquad \alpha=0.8.
\end{aligned}
\label{eq:hsp_probe_score}
\end{equation}
Here $\operatorname{Exp}$ is the spherical exponential map.
The means $\bar m,\bar G$ and positive standard deviations
$\sigma_m,\sigma_G$ are fitted on the ID-fit split and held fixed.
The probe keeps the initial centroid and uses no maximum over the trajectory.
Its score derivative is
\[
 \left.\frac{\partial s_\eta^{\rm probe}(x)}{\partial\eta}\right|_{\eta=0}
 =-\frac{\alpha}{\sigma_G}D(x).
\]
Thus $D$ measures the decrease
up to the common positive factor $\alpha/\sigma_G$. Define
\[
 M_\eta^{\rm probe}=s_\eta^{\rm probe}(X_{\rm in})-
 s_\eta^{\rm probe}(X_{\rm out}).
\]
The repulsive step gives
\begin{equation}
 M_\eta^{\rm probe}=M_0^{\rm probe}+\frac{\alpha\eta}{\sigma_G}
 \{D(X_{\rm out})-D(X_{\rm in})\}+O(\eta^2).
 \label{eq:hsp_pair_margin}
\end{equation}
Under the theorem's density conditions, the first-order AUROC change is
positive exactly when
$\mathbb E[D(X_{\rm out})-D(X_{\rm in})\mid M_0^{\rm probe}=0]>0$. The
uniform score remainder holds here because $m\in[0,1]$, $G\in[-1,1]$, and the
calibration scales are fixed. To summarize pairs near this boundary, let
$\mathcal I_\rho$ contain the empirical fraction $\rho$ of ID--OOD pairs
with the smallest absolute baseline margins, and define
\begin{equation}
 B_\rho=\frac{1}{|\mathcal I_\rho|}
 \sum_{(i,j)\in\mathcal I_\rho}
 \bigl[D(x_j^{\rm out})-D(x_i^{\rm in})\bigr].
 \label{eq:empirical_boundary_statistic}
\end{equation}
Here $x_i^{\rm in},x_j^{\rm out}$ are validation inputs. We report
$B_{5\%}$ and use $B_{100\%}$ for the mean over all pairs.
Writing $p_{M_0^{\rm probe}}$ for the baseline margin density, the theorem gives
\begin{equation}
\begin{aligned}
 \Delta\operatorname{AUROC}
 &=\eta\,p_{M_0^{\rm probe}}(0)\frac{\alpha}{\sigma_G}\\
 &\quad\times\mathbb E[D(X_{\rm out})-D(X_{\rm in})
       \mid M_0^{\rm probe}=0]+o(\eta).
\end{aligned}
\end{equation}
Both the boundary density and the calibration scale can vary across
settings. We therefore use $B_{5\%}$ as a diagnostic of direction, rather
than as a predictor of AUROC-change magnitude across settings.

\begin{figure*}[!t]
\centering
\includegraphics[width=\textwidth]{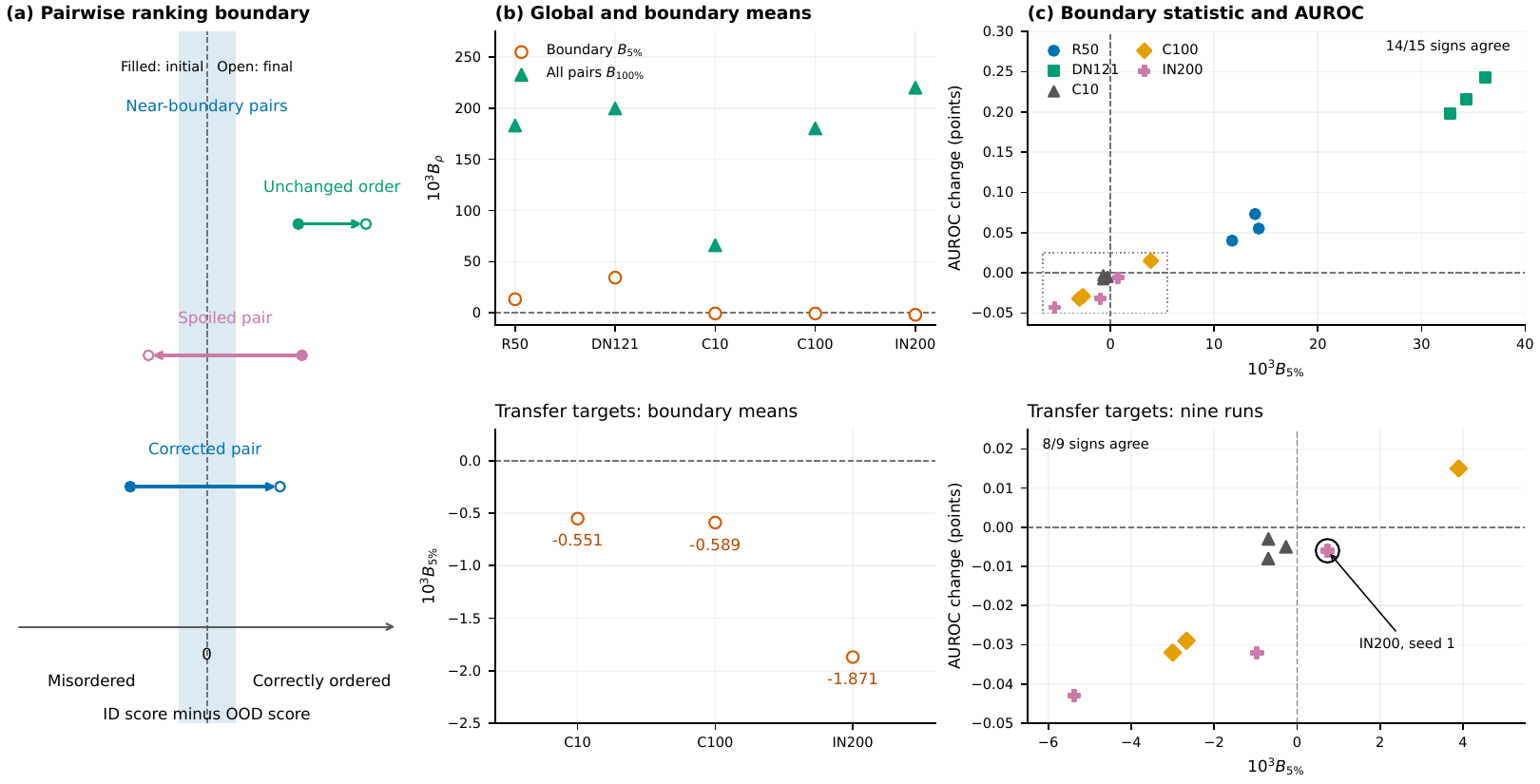}
\caption{Boundary-conditioned mechanism and validation evidence.
(a) AUROC changes when an ID--OOD pair crosses the ranking boundary;
filled and open markers denote initial and final margins.
(b) Mean $D_{\rm out}-D_{\rm in}$ over all pairs and over the closest $5\%$ of baseline
margins. The lower view enlarges the negative boundary means on the three
transfer targets. (c) Boundary statistic $B_{5\%}$ versus AUROC change at $\eta=.04$,
with the nine cross-dataset target--seed audits enlarged below.
Signs agree in 14 of 15 audits overall and 8 of 9 cross-dataset audits;
the circled point is the sole disagreement. AUROC changes are percentage
points. R50 and DN121 denote the ImageNet-1K ResNet-50 and DenseNet-121
settings; C10, C100, and IN200 denote CIFAR-10, CIFAR-100, and ImageNet-200.
Both statistics are computed on the same validation holdout.}
\label{fig:boundary_conditioned_stability}
\end{figure*}

The primary $\eta=0.04$ is selected on ImageNet-1K/ResNet-50 and then fixed
before all reported
transfers. Table~\ref{tab:hsp_transfer} shows small positive changes on both
ImageNet-1K backbones but no positive target mean. The $5\%$ boundary statistic
has the same sign as AUROC change in 14 of 15 runs across five
dataset--backbone settings. In contrast, the
unconditional statistic is positive on all three transfer targets and has the
opposite sign to their observed changes. Figure~\ref{fig:boundary_conditioned_stability}
visualizes this separation. This finite-window agreement is descriptive, not
a direct estimate of the infinitesimal derivative or a significance test.
The probe tests this local mechanism, not a performance-improvement claim;
detailed boundary fractions and descriptive correlations are in
\ref{app:stability_probe_audit}.

\begin{table*}[!t]
\centering
\caption{Stability probe with a fixed step size. Static and probe columns
report AUROC; $\overline B_{5\%}$ is $B_{5\%}$ from
Eq.~\eqref{eq:empirical_boundary_statistic}, averaged over three seeds.}
\label{tab:hsp_transfer}
\small
\setlength{\tabcolsep}{4.2pt}
\begin{tabular}{lccrrc}
\toprule
Target & Static & Probe & $\Delta$ & $10^3\overline B_{5\%}$ & Positive seeds \\
\midrule
IN-1K/ResNet-50    & $91.67\pm.45$ & $91.73\pm.44$ & $+.056$ & $+13.329$ & 3/3 \\
IN-1K/DenseNet-121 & $89.82\pm.40$ & $90.04\pm.38$ & $+.219$ & $+34.395$ & 3/3 \\
CIFAR-10           & $89.46\pm.81$ & $89.45\pm.81$ & $-.006$ & $-.551$ & 0/3 \\
CIFAR-100          & $84.96\pm.73$ & $84.95\pm.74$ & $-.016$ & $-.589$ & 1/3 \\
ImageNet-200       & $92.11\pm.32$ & $92.08\pm.33$ & $-.027$ & $-1.871$ & 0/3 \\
\bottomrule
\end{tabular}
\end{table*}

\subsection{Auditing the zero-step reduction}
\label{sec:The Static Limit}
\label{sec:Comprehensive OOD Detection Results}
\label{sec:Controlled Mechanism and Robustness Studies}
\label{sec:Computational Efficiency}

The static comparisons distinguish the effect of feature motion from the
choices used to construct the zero-step score. The six-stage comparison in
Table~\ref{tab:reduction_bridge_results} separates
field, scale, centroid, bank, and fusion effects on fixed ImageNet-1K
validation features. Candidate truncation is removed before S1, which already
uses the full class set; S2 then shares the class scales. These are additional
modeling choices beyond setting $T=0$. Sharing class scale produces the largest single increase;
radial centroid affinity and cosine then give identical rankings, exactly as
Proposition~\ref{prop:common_scale_equivalence} predicts. The empirical gain
from centroid replacement is not certified by the global approximation bound,
which preserves no audited pair by itself. Enlarging the bank and adding MSP
are empirical controls rather than consequences of the reduction.

\begin{table*}[!t]
\centering
\caption{Step-by-step static comparison on ImageNet-1K validation data.
Values are mean$\pm$standard deviation
over three detector banks; deltas are relative to MSP.}
\label{tab:reduction_bridge_results}
\small
\setlength{\tabcolsep}{5pt}
\begin{tabular}{llcccc}
\toprule
Stage & Interpretation & AUROC & FPR@95 & $\Delta$AUROC & $\Delta$FPR@95 \\
\midrule
MSP & output reference & $84.77\pm.00$ & $49.26\pm.00$ & $0.00$ & $0.00$ \\
S1 & exact full-class field & $73.16\pm.86$ & $72.34\pm2.61$ & $-11.61$ & $+23.08$ \\
S2 & common class scale & $87.06\pm.69$ & $46.44\pm.13$ & $+2.30$ & $-2.82$ \\
S3 & radial anchor centroid & $90.24\pm.61$ & $38.61\pm.75$ & $+5.47$ & $-10.65$ \\
S4 & cosine implementation & $90.24\pm.61$ & $38.61\pm.75$ & $+5.47$ & $-10.65$ \\
S5 & setup-bank centroid & $91.19\pm.67$ & $35.36\pm.34$ & $+6.43$ & $-13.90$ \\
S6 & fixed centroid--MSP control & $\mathbf{92.02\pm.52}$ & $\mathbf{31.58\pm.71}$ & $\mathbf{+7.26}$ & $\mathbf{-17.68}$ \\
\bottomrule
\end{tabular}
\end{table*}

The static controls are competitive but not uniformly dominant. CTM is
slightly stronger than the centroid--MSP baseline on four of eight
Near-/Far-OOD AUROC tasks, CADRef leads on
several ImageNet groups, and Scale is strongest on ImageNet-1K. Complete
baselines, component ablations, class-count analysis, bank/calibration
sensitivity, and batch-one efficiency are therefore placed in
\ref{app:complete_empirical_controls}. These comparisons show how
training-bank estimation and confidence fusion affect the static baselines.

\section{Discussion and Limitations}
\label{sec:Discussion}

\subsection{Interpretation of the mismatch theory}
\label{sec:The Surviving Geometric Structure}

The theory separates the driving field, the instantaneous detection score,
and the relative ordering of ID and OOD scores. The first-step theorem
connects the field to the instantaneous score for a step from rest. Keeping
the maximum score along a trajectory guarantees a nonnegative score increase,
but the exact pair decomposition shows why this need not improve AUROC.
For score families satisfying the additional expansion and density
conditions, the boundary theorem shows that the mean relative score
derivative at equal baseline scores, together with the margin density there,
determines the first-order AUROC change.

The transfer experiment shows that local score alignment and an unchanged
winning class are insufficient to ensure better ranking. This failure does
not require the destructive class interference allowed by the first-step
analysis. The separate probe illustrates why averaging score changes over
all pairs can miss the direction of AUROC change: the pairs near equal
baseline scores can move differently from the population as a whole.

The zero-step analysis identifies the geometric information present before
motion. Removing candidate truncation is an additional modeling choice.
The approximation bound then controls the replacement of full-class anchor
fields by radial centroid scores. Under shared scales, the resulting
centroid score is rank-equivalent to maximum cosine similarity. CTM is an
existing full-bank, feature-norm-weighted member of this centroid family.
The step-by-step experiment examines these scoring choices separately from
the additional effects of bank size and confidence fusion.
These comparisons help attribute static performance to scoring choices;
they do not establish that centroid replacement improves AUROC in general.

\subsection{A practical diagnostic and its scope}
\label{sec:practical_diagnostic}

Theorem~\ref{thm:boundary_auroc_derivative} suggests a validation diagnostic
for score refinements satisfying its regularity conditions. On
ID and proxy-OOD validation data allowed by the evaluation protocol, estimate
each sample's score derivative, form ID--OOD pairs, retain a fixed fraction
with the smallest absolute baseline margins, and average the ID-minus-OOD
relative derivative. Its sign provides a local prediction to compare with
the observed validation AUROC change. Stability across
seeds and boundary fractions should be checked in the intended validation
setting. The mean relative derivative over all pairs provides a useful
comparison, because its sign can differ from that near the ranking boundary.

The empirical conclusions are limited to the tested settings. The dynamics
use fixed initial candidate sets, and the no-switch audit covers $T\le10$;
the ImageNet audit excludes classes outside the retained top-20 set. The
score-aligned transfers and the separate probe use validation data. The
probe's 15 runs cover three seeds in each of five dataset--backbone settings,
with shared validation images within each setting. For each ImageNet-1K
backbone, variation is over sampled banks. These runs provide descriptive
evidence, not 15 independent datasets or a significance test.

The diagnostic also has a local scope. The fixed $5\%$ window summarizes
pairs near the ranking boundary; it does not directly estimate the
conditional mean at exactly equal scores. The observed AUROC changes at
$\eta=0.04$ are finite-step changes, several of which are very small.
Agreement between their signs and the boundary statistic therefore supports
a local interpretation without guaranteeing gains at other step sizes or
on other distributions. Theorem~\ref{thm:no_universal_auc_gain} further shows
that changing a strict baseline ordering can reduce AUROC for some ID/OOD
distributions.

The numerical model and static approximation impose further limits. The
original position-normalized scheme has substantial energy and tangency
errors, so the geodesic control provides the evidence for constrained
feature motion. The centroid approximation bound is too loose to certify
the observed static rankings. Whether the diagnostic extends to transformers
and foundation-model features, or can be used with ID-only validation
objectives, remains open.

\balance
\section{Conclusion}

Class attraction and local score alignment do not guarantee improved OOD
ranking. The transfer study demonstrates this failure even without observed
winning-class changes within the fixed candidate sets. Under regularity
conditions, the mean relative score derivative at the ID--OOD ranking
boundary determines the direction of first-order AUROC change. A separate
perturbation study finds that near-boundary relative score changes generally
track the direction of small AUROC changes. Through centroid approximation
and shared-scale ranking equivalence, the static analysis relates zero-step
anchor scores to a centroid family that includes CTM. Together, these
results show why samplewise score increases are insufficient and why feature
refinement should be evaluated by its effect on ID--OOD ordering.

\FloatBarrier
\section*{Code availability}
Implementation and evaluation scripts are available in our public
{\color{blue}\href{https://github.com/zilamwong2005-math/Hamiltonian-Centroid-OOD}
{GitHub repository}}. Software versions, pinned dependencies, and checkpoint
identifiers are specified in the appendix.
\clearpage
\onecolumn
\appendix
\renewcommand{\thetheorem}{\Alph{section}.\arabic{theorem}}
\setlength{\intextsep}{8pt plus 2pt minus 2pt}
\setlength{\textfloatsep}{8pt plus 2pt minus 2pt}
\section{Mathematical Details and Supplementary Results}
\label{app:supplementary}

\subsection{Fundamental-model properties and numerical update}
\label{app:fundamental_model_details}
\label{app:geometric_proofs}

\begin{lemma}[Transport of local class concentration]
\label{lem:concentration_transport}
Let $\mathcal M_c\subset\mathcal X$ be the class-$c$ image manifold and let
$q:\mathcal X\to\mathbb S^{d-1}$ be $L_c$-Lipschitz on
$U_c=\{x:d_{\mathcal X}(x,\mathcal M_c)\le r_c\}$. If a class-$c$ random
image $X_c$ satisfies
\begin{equation}
 \Pr\!\left[d_{\mathcal X}(X_c,\mathcal M_c)\le r_c\right]\ge 1-\delta_c,
 \label{eq:appendix_input_concentration}
\end{equation}
then its normalized feature satisfies
\begin{equation}
 \Pr\!\left[d_{\mathbb S}\!\left(q(X_c),q(\mathcal M_c)\right)
 \le L_c r_c\right]\ge 1-\delta_c,
 \label{eq:appendix_feature_concentration}
\end{equation}
where $d_{\mathbb S}(u,B)=\inf_{v\in B}\lVert u-v\rVert_2$.
\end{lemma}

\begin{proof}
For every $x\in U_c$ and $m\in\mathcal M_c$, Lipschitz continuity gives
\[
 d_{\mathbb S}\!\left(q(x),q(\mathcal M_c)\right)
 \le \lVert q(x)-q(m)\rVert_2
 \le L_c d_{\mathcal X}(x,m).
\]
Taking the infimum over $m\in\mathcal M_c$ yields
$d_{\mathbb S}(q(x),q(\mathcal M_c))\le
L_c d_{\mathcal X}(x,\mathcal M_c)$. Applying this deterministic inequality
on the event in Eq.~\eqref{eq:appendix_input_concentration} gives
Eq.~\eqref{eq:appendix_feature_concentration}.
\end{proof}

\paragraph{Original position-normalized update.}
The original trajectory experiments use ambient kicks with position
normalization. With the initial candidate set fixed, one update is
\begin{equation}
 \begin{aligned}
 p_{n+\frac12}&=p_n+\frac{\Delta t}{2}\nabla_qB(q_n;q_0),\\
 \widetilde q_{n+1}&=q_n+\Delta t\,p_{n+\frac12},\\
 q_{n+1}&=\widetilde q_{n+1}/\lVert\widetilde q_{n+1}\rVert_2,\\
 p_{n+1}&=p_{n+\frac12}+\frac{\Delta t}{2}\nabla_qB(q_{n+1};q_0).
 \end{aligned}
 \label{eq:projected_leapfrog}
\end{equation}
The normalization keeps positions on $\mathbb S^{d-1}$, but momentum is not
constrained to the tangent space. This is a distinct heuristic discretization,
not a symplectic integrator for Eq.~\eqref{eq:hamilton_equations}; no long-time
energy or tangency guarantee is claimed for it.

\paragraph{Constrained geodesic update.}
For the score-aligned audit, let $F$ denote the aggregate or single-class
driver in Section~\ref{sec:trajectory_ranking}. Starting with tangent $p_n$,
set
\[
 \widetilde v_n=p_n+\tfrac12\Delta t P_{q_n}\nabla F(q_n),
 \qquad \nu_n=\lVert\widetilde v_n\rVert_2.
\]
Exact kinetic drift followed by a tangent kick gives
\begin{equation}
\begin{aligned}
 q_{n+1}&=\cos(\Delta t\nu_n)q_n+
 \frac{\sin(\Delta t\nu_n)}{\nu_n}\widetilde v_n,\\
 \widetilde p_{n+1}&=\cos(\Delta t\nu_n)\widetilde v_n-
 \nu_n\sin(\Delta t\nu_n)q_n,\\
 p_{n+1}&=\widetilde p_{n+1}+
 \tfrac12\Delta t P_{q_{n+1}}\nabla F(q_{n+1}).
\end{aligned}
\label{eq:geodesic_kdk}
\end{equation}
The quotient is interpreted as $\Delta t$ when $\nu_n=0$. For a tangent
vector $v$, the spherical exponential map is
\[
 \operatorname{Exp}_q(v)=\cos(\lVert v\rVert_2)q+
 \frac{\sin(\lVert v\rVert_2)}{\lVert v\rVert_2}v,
 \qquad \operatorname{Exp}_q(0)=q.
\]
The two kicks and
the drift preserve the tangent bundle in exact arithmetic. A deterministic
active class is selected at a current-winner tie, and smooth-flow claims are
made only away from ties. Both schemes keep $\mathcal J(q_0)$ fixed; only the
active class within that set may change.

\subsection{Proofs for the attraction--ranking analysis}
\label{app:attraction_ranking_proofs}

\begin{proof}[Proof of Theorem~\ref{thm:first_step_mismatch}]
The first half-step and position update give
\[
 p_{1/2}=\frac{\Delta t}{2}\nabla B(q_0;q_0),
 \qquad
 \widetilde q_1=q_0+\frac{\Delta t^2}{2}\nabla B(q_0;q_0).
\]
For a unit vector $q$ and small $h$,
\[
 \frac{q+hv}{\lVert q+hv\rVert_2}
 =q+h(I-qq^\top)v+O(h^2).
\]
Substitution of $h=\Delta t^2/2$ proves
Eq.~\eqref{eq:first_step_expansion} for the original update. For the geodesic
scheme, $\nu_0=\tfrac12\Delta t\lVert b_0\rVert_2$ and
$q_1=\operatorname{Exp}_{q_0}(\tfrac12\Delta t^2b_0)$, which has the same
expansion with remainder $O(\Delta t^4)$. Taylor expansion at $q_0$ gives
\[
 B(q_1;q_0)-B(q_0;q_0)
 =\frac{\Delta t^2}{2}\nabla B(q_0;q_0)^\top
 P_{q_0}\nabla B(q_0;q_0)+O(\Delta t^4),
\]
which is Eq.~\eqref{eq:total_field_local_gain} because $P_{q_0}$ is an
orthogonal projector. The same expansion for $A_c$ yields
Eq.~\eqref{eq:class_field_local_gain}.
\end{proof}

\begin{proof}[Proof of Proposition~\ref{prop:fixed_winner_equivalence}]
Write $z_n^{\rm init}=(q_n^{\rm init},p_n^{\rm init})$ and
$z_n^{\rm cur}=(q_n^{\rm cur},p_n^{\rm cur})$. The initial states agree.
Suppose $z_n^{\rm init}=z_n^{\rm cur}=z_n$. Since $c_0$ is the unique
maximizer at $q_n$, continuity gives a neighborhood on which
$F_{\rm cur}=F_{\rm init}=KA_{c_0}$. Hence
\[
 P_{q_n}\nabla F_{\rm cur}(q_n)
 =P_{q_n}\nabla F_{\rm init}(q_n).
\]
The first kicks produce identical intermediate momenta, and the deterministic
drift produces the same $q_{n+1}$ and intermediate momentum. The unique-winner
assumption at $q_{n+1}$ gives the same equality of projected gradients there,
so the second kicks imply $z_{n+1}^{\rm init}=z_{n+1}^{\rm cur}$.
Induction proves equality of all states and active classes. The same argument
applies at any additional internal force evaluation points covered by the
assumption. Finally, at every retained state,
\[
 m_{\mathcal J}(q_n;q_0)=A_{c_0}(q_n),\qquad
 F_{\rm cur}(q_n)=F_{\rm init}(q_n).
\]
Taking maxima over $0\le n\le T$ proves equality of trajectory scores.
\end{proof}

\begin{proof}[Proof of Corollary~\ref{cor:gradient_interference}]
Linearity of $P_{q_0}$ and Eq.~\eqref{eq:candidate_field} imply
\[
 P_{q_0}\nabla B(q_0;q_0)=K\sum_{c\in\mathcal J(q_0)}g_c.
\]
Hence
\[
 \left\langle g_{c^\star},P_{q_0}\nabla B(q_0;q_0)\right\rangle
 =K\lVert g_{c^\star}\rVert_2^2[1+I(q_0)],
\]
which proves Eq.~\eqref{eq:interference_expansion}. If $I(q_0)<-1$, its
leading coefficient is negative. It also follows that
$P_{q_0}\nabla B(q_0;q_0)\ne0$, so
Eq.~\eqref{eq:total_field_local_gain} has a positive leading coefficient.
The condition $I(q_0)<-1$ requires at least two candidate classes. Since
$c^\star$ is the unique maximizer, define
\[
 \delta_*=A_{c^\star}(q_0)-
 \max_{c\in\mathcal J(q_0)\setminus\{c^\star\}}A_c(q_0)>0.
\]
Continuity and finiteness of the candidate set give a neighborhood $U$ of
$q_0$ on which $|A_c(q)-A_c(q_0)|<\delta_*/3$ for every candidate class.
Thus $A_{c^\star}(q)-A_c(q)>\delta_*/3$ for all $q\in U$ and $c\ne c^\star$.
Since $q_1\to q_0$ as $\Delta t\to0$, the maximizer is unchanged for all
sufficiently small nonzero steps. The instantaneous maximum therefore
decreases while $B$ increases.
\end{proof}

\begin{proof}[Proof of Theorem~\ref{thm:auroc_lift_decomposition}]
Under the no-tie assumption,
\[
 \operatorname{AUROC}_T=\Pr(M_0+D_T>0),
 \qquad
 \operatorname{AUROC}_0=\Pr(M_0>0).
\]
The pointwise indicator identity is
\[
 \mathbf1_{\{M_0+D_T>0\}}-\mathbf1_{\{M_0>0\}}
 =\mathbf1_{\mathcal C_T}-\mathbf1_{\mathcal S_T}.
\]
Taking expectations proves the first statement. A changed ordering requires
$|D_T|\ge|M_0|$, so
\[
 |\Delta\operatorname{AUROC}|
 \le\mathbb E|\mathbf1_{\mathcal C_T}-\mathbf1_{\mathcal S_T}|
 \le\Pr(|M_0|\le|D_T|).
\]
If ties are present, Eq.~\eqref{eq:auroc_convention} instead gives the exact
identity
\begin{equation}
\begin{aligned}
 \Delta\operatorname{AUROC}
 &=\Pr(\mathcal C_T)-\Pr(\mathcal S_T)\\
 &\quad+\tfrac12\bigl[\Pr(M_0+D_T=0)-\Pr(M_0=0)\bigr].
\end{aligned}
\label{eq:auroc_tie_correction}
\end{equation}
Thus empirical pair counts must include the change in tie mass when compared
with AUROC computed using half credit for ties.
\end{proof}

\begin{proof}[Proof of Theorem~\ref{thm:boundary_auroc_derivative}]
First control the tail of the relative score derivative. Omit the remainder
and write $Z_\eta=M_0+\eta U$. Set $a_\eta=\delta/(2|\eta|)$ for
$\eta\ne0$. Since $\mathbb E|U|<\infty$,
\[
 \frac{\Pr(|U|>a_\eta)}{|\eta|}
 \le \frac{2}{\delta}
 \mathbb E\!\left[|U|\mathbf 1_{\{|U|>a_\eta\}}\right]
 \longrightarrow0.
\]
Next evaluate the derivative on the remaining region. For $|u|\le a_\eta$,
the oriented integral identity
\[
 \frac{1}{\eta}\int_{-\eta u}^{0}p_{M_0,U}(m,u)\,dm
 =u\int_0^1 p_{M_0,U}(-t\eta u,u)\,dt
\]
is bounded in absolute value by $|u|g(u)$. Continuity in $m$, dominated
convergence, and the negligible tail therefore give
\[
 \lim_{\eta\to0}
 \frac{\Pr(Z_\eta>0)-\Pr(M_0>0)}{\eta}
 =\int u\,p_{M_0,U}(0,u)\,du.
\]

Finally restore the remainder. Since
\[
 |R_\eta(X_{\rm in})-R_\eta(X_{\rm out})|
 \le2\varepsilon_\eta
 \quad\text{almost surely},
\]
a change in ordering,
including the creation of a tie, requires $|Z_\eta|\le2\varepsilon_\eta$.
For sufficiently small $|\eta|$, the corresponding intervals in $m$ lie in
$[-\delta,\delta]$ whenever $|u|\le a_\eta$, so
\[
 \Pr(|Z_\eta|\le2\varepsilon_\eta)
 \le4\varepsilon_\eta\int g(u)\,du+\Pr(|U|>a_\eta)
 =o(|\eta|).
\]
Thus the remainder does not change the derivative, also under the usual
half-credit convention for ties. Finally,
$\int u\,p_{M_0,U}(0,u)\,du=p_{M_0}(0)\mathbb E[U\mid M_0=0]$ by the stated density
definition.
\end{proof}

\paragraph{Connection to a single trajectory step.}
For a unique initial winner, Theorem~\ref{thm:first_step_mismatch} and the
positive class margin imply, with $\eta=\Delta t^2/2\ge0$,
\[
 m_{\mathcal J}(q_1;q_0)-m_{\mathcal J}(q_0;q_0)
 =\eta\langle g_{c^\star},b_0\rangle+O(\eta^2).
\]
Since $u\mapsto\max\{u,0\}$ is $1$-Lipschitz, retaining the initial score gives
\[
 S_1(q_0)-S_0(q_0)
 =\eta\max\{\langle g_{c^\star},b_0\rangle,0\}+O(\eta^2).
\]
This is a pointwise expansion. Applying the boundary theorem to this family
would additionally require a uniform remainder over the two populations
and the stated joint-density conditions. The separate probe in
Section~\ref{sec:stability_probe_transfer} provides a simpler score family
with a uniform remainder.

\subsection{Proofs for the static reduction}
\label{app:static_reduction_proofs}

\begin{proof}[Proof of Proposition~\ref{prop:centroid_approximation}]
The definitions give
\[
 A_c(q)=W_c\sum_j\pi_{cj}\phi\!\left(
 \frac{\lVert q-a_{cj}\rVert_2}{\sigma_c}\right).
\]
The reverse triangle inequality and Lipschitz continuity of $\phi$ imply
\[
\begin{aligned}
&\left|A_c(q)-W_c\phi\!\left(
 \frac{\lVert q-\bar a_c\rVert_2}{\sigma_c}\right)\right|\\
&\quad\le \frac{W_cL_\phi}{\sigma_c}
\sum_j\pi_{cj}\left|
 \lVert q-a_{cj}\rVert_2-\lVert q-\bar a_c\rVert_2\right|\\
&\quad\le \frac{W_cL_\phi}{\sigma_c}
\sum_j\pi_{cj}\lVert a_{cj}-\bar a_c\rVert_2.
\end{aligned}
\]
Because $\bar a_c=\lVert\bar a_c\rVert_2\mu_c$ and the anchors lie on the
unit sphere,
\[
 \lVert\bar a_c-\mu_c\rVert_2
 =\left|1-\lVert\bar a_c\rVert_2\right|.
\]
A second application of the Lipschitz inequality bounds the replacement of
$\bar a_c$ by $\mu_c$ by
$W_cL_\phi\left|1-\lVert\bar a_c\rVert_2\right|/\sigma_c$. The triangle
inequality combines the two displays into
Eq.~\eqref{eq:spherical_centroid_bound}. To obtain the score bound, use
\[
 \left|\max_c A_c(q)-\max_c\widetilde A_c(q)\right|
 \le\max_c\left|A_c(q)-\widetilde A_c(q)\right|
\]
and apply the class-wise bound.
For two queries $q,q'$, it follows that
\[
 \widetilde S_0(q)-\widetilde S_0(q')
 \ge S_0^{\rm all}(q)-S_0^{\rm all}(q')-2\epsilon.
\]
This proves the margin condition in Eq.~\eqref{eq:centroid_margin_preservation}.
\end{proof}

\begin{proof}[Proof of Proposition~\ref{prop:common_scale_equivalence}]
For unit vectors,
\[
 \lVert q-\mu_c\rVert_2^2=2-2q^\top\mu_c.
\]
Thus maximizing cosine similarity over classes minimizes radial distance.
Because $\phi$ is strictly decreasing and $W>0$, substituting the minimizing
distance gives Eq.~\eqref{eq:radial_cosine_score_equivalence}. The map
$g\mapsto W\phi(\sqrt{2-2g}/\sigma)$ is strictly increasing on $[-1,1]$,
which proves both the sample-ranking and class-selection statements.
\end{proof}

\begin{proof}[Proof of Proposition~\ref{prop:ctm_special_case}]
CTM forms an unnormalized class mean proportional to
\[
 \sum_{i\in\mathcal B_c} f(x_i)
 =\sum_{i\in\mathcal B_c}r_iq(x_i).
\]
Normalizing this vector removes the irrelevant factor $|\mathcal B_c|^{-1}$
and yields $\mu_c(\mathcal B_{\rm all},r)$ in
Eq.~\eqref{eq:general_centroid_family}. Maximizing its cosine with $q(x)$ is
therefore exactly $G_{\mathcal B_{\rm all},r}(x)$.
\end{proof}

\subsection{Score-aligned transfer and winner-switch audit}
\label{app:aligned_dynamics_audit}

Table~\ref{tab:aligned_transfer_detail} gives the per-seed validation results
under the locked geodesic $T=5$ rule. Corrected and spoiled columns are the
probability masses from Theorem~\ref{thm:auroc_lift_decomposition}, expressed
as percentages. AUROC uses half credit for ties; when tie mass changes, the
exact relation includes Eq.~\eqref{eq:auroc_tie_correction}. Displayed pair
masses and AUROC changes are rounded separately.

\begin{table}[H]
\centering
\caption{Per-seed score-aligned mechanism audit. Static, aggregate, and aligned
columns report AUROC; C/S gives corrected/spoiled pair mass in percent for the
aligned flow relative to $T=0$.}
\label{tab:aligned_transfer_detail}
\small
\setlength{\tabcolsep}{4.5pt}
\begin{tabular}{lrrrrr}
\toprule
Validation benchmark & Seed & $T=0$ & Aggregate & Aligned & C/S \\
\midrule
ImageNet-1K & 0 & 72.27 & 65.77 & 73.07 & 2.74/1.94 \\
            & 1 & 70.84 & 66.24 & 72.30 & 3.02/1.57 \\
            & 2 & 72.27 & 65.95 & 73.41 & 3.02/1.88 \\
\midrule
CIFAR-10    & 0 & 89.23 & 89.93 & 87.63 & 0.57/2.17 \\
            & 1 & 89.18 & 90.18 & 87.93 & 0.77/2.01 \\
            & 2 & 89.85 & 90.29 & 88.39 & 0.46/1.92 \\
CIFAR-100   & 0 & 85.15 & 86.44 & 84.73 & 0.50/0.91 \\
            & 1 & 83.40 & 84.40 & 82.95 & 0.53/0.97 \\
            & 2 & 84.24 & 85.31 & 83.84 & 0.48/0.88 \\
ImageNet-200& 0 & 87.62 & 85.79 & 87.16 & 0.78/1.25 \\
            & 1 & 88.88 & 88.44 & 88.59 & 0.75/1.04 \\
            & 2 & 87.77 & 85.98 & 86.85 & 0.59/1.51 \\
\bottomrule
\end{tabular}
\end{table}

The post-hoc switch audit uses all retained states $T=0,\ldots,10$ rather than
only the selected $T=5$ result. Table~\ref{tab:winner_switch_audit} aggregates the
three seeds. ``Candidate classes'' denotes the set fixed at initialization; the
ImageNet-200 and CIFAR-100 statement does not claim that classes outside the
top-20 set were re-evaluated during motion. The two driver implementations
also have zero maximum absolute difference in trajectory score,
instantaneous score, driver value, and Hamiltonian energy.

\begin{table}[H]
\centering
\caption{Winner-switch audit over 20,205 sample--seed trajectories and 11
retained states. Switch counts are summed over seeds and populations.}
\label{tab:winner_switch_audit}
\small
\setlength{\tabcolsep}{6pt}
\begin{tabular}{lrrrrr}
\toprule
Benchmark & Candidates & ID & OOD & Current switches & Initial switches \\
\midrule
CIFAR-10     & all 10 & 3000 & 3000 & 0 & 0 \\
CIFAR-100    & top 20 & 3000 & 2916 & 0 & 0 \\
ImageNet-200 & top 20 & 3000 & 5289 & 0 & 0 \\
\bottomrule
\end{tabular}
\end{table}

The mechanism and numerical checks average over ID/OOD populations and three
anchor banks. The aggregate and current-winner alignment cosines are $0.634$
and $1.000$; neither produces a first-step winner decrease, and the audited
fraction satisfying $I(q_0)<-1$ is zero. Thus these data do not demonstrate
destructive class interference, although they test its local alignment
expansion. Both attain $100\%$ Taylor sign agreement. Winner and driver Taylor residuals have
log--log slopes $4.138$--$4.162$ and $3.833$--$4.179$, consistent with the
$O(\Delta t^4)$ remainder underlying
Definition~\ref{def:driver_score_alignment}. The
geodesic controls keep maximum sphere error below $2.38\times10^{-7}$ and
maximum tangency error below $1.49\times10^{-7}$; their mean maximum relative
energy drifts are $0.0352$ (aggregate) and $0.00265$ (current winner). For
comparison, the original retraction-based aggregate integrator has drift
$1.327$ and tangency error $8.604$. The stable geodesic experiments therefore
provide the basis for the aligned-flow mechanism claim; their failed locked
transfer is not explained by leaving the sphere or tangent bundle.

\subsection{Boundary-conditioned stability-probe audit}
\label{app:stability_probe_audit}

The primary stability-probe rule fixes $\eta=0.04$ from the ImageNet-1K
ResNet-50 selection split. No transfer target re-selects the step size. A
secondary scale control sets
\[
 \eta_{\rm ID}=0.0025/\mathbb E_{\rm ID,fit}[D]
\]
using only the ID fit split; its results are retained in the accompanying CSV
archive as an additional control rather than in the primary tables. Both
rules retain the same MSP component and the same ID-fit standardization as the
static score.

The probe in Eq.~\eqref{eq:hsp_probe_score} is generated by the conditional
potential
$V_x(q)=\gamma(1-m(x))q^\top\mu_{\hat c(x)}$, where $\gamma>0$ and the initial confidence
and centroid are held fixed. One geodesic kick--drift step from rest has
$\eta=\gamma\Delta t^2/2$; the final kick does not change its position. Thus
this is a one-step repulsive perturbation, not an exact finite-time
solution of the full nonlinear flow or a maximum score along a trajectory. For
completeness, with
$\theta=\eta(1-m)\sqrt{1-G^2}$, the initial-winning geodesic has the exact
endpoint
\[
 G_\eta=G\cos\theta-\sqrt{1-G^2}\sin\theta
       =G-\eta(1-m)(1-G^2)+O(\eta^2).
\]
Substitution into the fixed standardized fusion yields
Eq.~\eqref{eq:hsp_pair_margin} in the main text.
Table~\ref{tab:hsp_per_seed_detail} reports the primary rule.
Corrected and spoiled masses are percentages of all ID--OOD pairs. Their
difference gives the AUROC change in percentage points, with the tie
adjustment in Eq.~\eqref{eq:auroc_tie_correction} when needed.

\begin{table}[H]
\centering
\caption{Per-seed stability-probe results. $B_{5\%}$ is the mean
$D_{\rm out}-D_{\rm in}$ among the nearest $5\%$ of pairs to the static
ranking boundary. C/S denotes corrected/spoiled pair mass in percent.}
\label{tab:hsp_per_seed_detail}
\scriptsize
\setlength{\tabcolsep}{3.1pt}
\begin{tabular}{lrrrrrr}
\toprule
Validation target & Seed & Static & Probe & $\Delta$ & $10^3 B_{5\%}$ & C/S \\
\midrule
IN-1K / R50 & 0 & 92.165 & 92.205 & $+0.040$ & $+11.734$ & 0.296/0.256 \\
             & 1 & 91.269 & 91.342 & $+0.073$ & $+13.950$ & 0.336/0.263 \\
             & 2 & 91.583 & 91.638 & $+0.055$ & $+14.302$ & 0.325/0.269 \\
IN-1K / DN121 & 0 & 90.065 & 90.281 & $+0.216$ & $+34.302$ & 0.553/0.337 \\
               & 1 & 89.357 & 89.600 & $+0.243$ & $+36.159$ & 0.581/0.338 \\
               & 2 & 90.049 & 90.246 & $+0.198$ & $+32.726$ & 0.537/0.339 \\
\midrule
CIFAR-10 & 0 & 88.859 & 88.851 & $-0.008$ & $-0.694$ & 0.008/0.016 \\
         & 1 & 89.135 & 89.131 & $-0.003$ & $-0.691$ & 0.006/0.010 \\
         & 2 & 90.377 & 90.372 & $-0.005$ & $-0.268$ & 0.001/0.006 \\
CIFAR-100& 0 & 85.375 & 85.343 & $-0.032$ & $-2.998$ & 0.108/0.140 \\
         & 1 & 84.127 & 84.098 & $-0.029$ & $-2.666$ & 0.122/0.152 \\
         & 2 & 85.392 & 85.407 & $+0.015$ & $+3.897$ & 0.143/0.128 \\
IN-200   & 0 & 92.069 & 92.026 & $-0.043$ & $-5.378$ & 0.106/0.149 \\
         & 1 & 92.448 & 92.441 & $-0.006$ & $+0.733$ & 0.113/0.120 \\
         & 2 & 91.819 & 91.788 & $-0.032$ & $-0.966$ & 0.109/0.141 \\
\bottomrule
\end{tabular}
\end{table}

The single sign disagreement occurs for ImageNet-200 seed~1, where both the
boundary statistic and the AUROC change are close to zero. The pair
decomposition accounts for AUROC change by construction. The additional
finding is the sign agreement with $B_{5\%}$, which is computed from
baseline margins and score derivatives on the same holdout.
Across the 15 runs in five dataset--backbone settings, the descriptive
Pearson and Spearman correlations between $B_{5\%}$ and AUROC change are
$0.991$ and $0.979$. These pooled correlations do not establish a common
quantitative relation across settings: the boundary density and calibration
scale also enter the first-order AUROC change. Runs within a setting share
validation images, so these correlations are not significance tests based
on 15 independent datasets.

Table~\ref{tab:hsp_boundary_fractions} varies the width of the empirical
boundary layer. ImageNet-1K has a positive boundary statistic throughout. For every transfer
target the full-pair mean is positive, yet the narrowest layers are negative.
This is the observed global-to-boundary sign reversal behind the failed
transfer.

\begin{table}[H]
\centering
\caption{Mean boundary statistic $B_\rho$, multiplied by $10^3$, over
three seeds. Columns retain the stated fraction of ID--OOD pairs with the
smallest absolute static margin.}
\label{tab:hsp_boundary_fractions}
\scriptsize
\setlength{\tabcolsep}{4.2pt}
\begin{tabular}{lrrrrrr}
\toprule
Validation target & 1\% & 2.5\% & 5\% & 10\% & 25\% & 100\% \\
\midrule
IN-1K / R50   & 11.128 & 12.013 & 13.329 & 18.480 & 43.652 & 182.907 \\
IN-1K / DN121 & 32.548 & 33.590 & 34.395 & 38.253 & 59.390 & 199.719 \\
CIFAR-10      & $-0.527$ & $-0.567$ & $-0.551$ & $-0.457$ & 0.277 & 65.996 \\
CIFAR-100     & $-0.934$ & $-1.616$ & $-0.589$ & 1.831 & 16.693 & 180.182 \\
IN-200        & $-4.551$ & $-3.871$ & $-1.871$ & 4.976 & 40.295 & 219.853 \\
\bottomrule
\end{tabular}
\end{table}

The predefined transfer criteria for AUROC and boundary consistency are not
met, although the FPR@95 criterion is met. The probe was therefore not
evaluated on the Near-/Far-OOD test sets; no corresponding test loader or
score file was accessed in this study.

\subsection{Full ImageNet trajectory ablation}

Table~\ref{tab:trajectory_full} reports the aggregated ImageNet trajectory grid.
Values are Near-/Far-OOD macro
averages; each cell is AUROC/FPR@95. The Gaussian and IMQ fields share the same
anchors, optimizer schedule, candidate count, and step size.

\begin{table}[H]
\centering
\caption{ImageNet trajectory ablation. ImageNet-200 varies three official
model seeds; ImageNet-1K varies three centroid banks on one fixed backbone.}
\label{tab:trajectory_full}
\small
\setlength{\tabcolsep}{5pt}
\begin{tabular}{llrrrr}
\toprule
ID & Potential & $T=0$ & $T=1$ & $T=3$ & $T=10$ \\
\midrule
IN-200 Near & Gaussian & 79.44/64.42 & 79.64/63.99 & 80.22/61.89 & 79.63/62.10 \\
IN-200 Far  & Gaussian & 91.18/31.66 & 91.29/31.36 & 90.68/33.10 & 88.82/34.61 \\
IN-200 Near & IMQ      & 78.23/65.27 & 78.41/64.95 & 79.34/62.99 & 78.16/63.37 \\
IN-200 Far  & IMQ      & 89.79/34.33 & 89.90/34.06 & 90.17/33.50 & 87.09/37.39 \\
\midrule
IN-1K Near & Gaussian & 66.14/77.84 & 64.51/79.27 & 61.42/79.63 & 60.94/79.63 \\
IN-1K Far  & Gaussian & 77.44/65.16 & 72.71/71.29 & 74.37/69.75 & 68.51/70.71 \\
IN-1K Near & IMQ      & 63.46/80.45 & 60.91/82.52 & 61.94/81.33 & 58.03/82.91 \\
IN-1K Far  & IMQ      & 73.76/69.48 & 65.68/78.21 & 75.58/66.79 & 63.44/75.60 \\
\bottomrule
\end{tabular}
\end{table}

Table~\ref{tab:imagenet1k_t10_detail} gives the uncertainty and precision--recall
statistics for the ImageNet-1K $T=10$ column.

\begin{table}[H]
\centering
\caption{ImageNet-1K $T=10$ full-evaluation results over three independently
sampled centroid banks on the fixed ResNet-50. Values are mean$\pm$standard
deviation; ID accuracy is $76.1867\%$ for every run.}
\label{tab:imagenet1k_t10_detail}
\scriptsize
\begin{tabular}{llrrrr}
\toprule
Potential & Group & AUROC & FPR@95 & AUPR-IN & AUPR-OUT \\
\midrule
Gaussian & Near & $60.94\pm.07$ & $79.63\pm1.29$ & $75.14\pm.17$ & $34.70\pm.24$ \\
Gaussian & Far  & $68.51\pm.33$ & $70.71\pm1.75$ & $90.86\pm.26$ & $26.15\pm.04$ \\
IMQ & Near & $58.03\pm.83$ & $82.91\pm1.79$ & $73.52\pm.22$ & $33.27\pm.14$ \\
IMQ & Far  & $63.44\pm1.05$ & $75.60\pm1.52$ & $88.96\pm.53$ & $23.21\pm.15$ \\
\bottomrule
\end{tabular}
\end{table}

\begin{table}[H]
\centering
\caption{Five potential families on CIFAR. Bandwidths are fitted from
zero-step class affinities and scores use $T=120$. Values are macro averages
over all six OOD datasets and three seeds.}
\label{tab:potential_ablation}
\small
\setlength{\tabcolsep}{4pt}
\begin{tabular}{llcc}
\toprule
ID & Potential & AUROC & FPR@95 \\
\midrule
CIFAR-10 & Gaussian  & $89.77\pm0.13$ & $35.18\pm1.24$ \\
 & Laplacian & $\mathbf{90.97\pm0.36}$ & $32.82\pm2.19$ \\
 & Cauchy    & $89.94\pm0.24$ & $34.55\pm1.11$ \\
 & IMQ       & $90.66\pm0.05$ & $\mathbf{31.73\pm1.27}$ \\
 & Mat\'ern-$3/2$ & $89.97\pm0.44$ & $37.46\pm3.18$ \\
\midrule
CIFAR-100 & Gaussian  & $\mathbf{82.58\pm0.61}$ & $\mathbf{53.70\pm0.91}$ \\
 & Laplacian & $81.60\pm0.46$ & $55.16\pm0.94$ \\
 & Cauchy    & $81.19\pm0.36$ & $54.99\pm0.81$ \\
 & IMQ       & $82.35\pm0.53$ & $54.44\pm0.73$ \\
 & Mat\'ern-$3/2$ & $81.07\pm0.74$ & $54.46\pm0.92$ \\
\bottomrule
\end{tabular}
\end{table}

\subsection{Effective-rank anchor-weighting ablation}

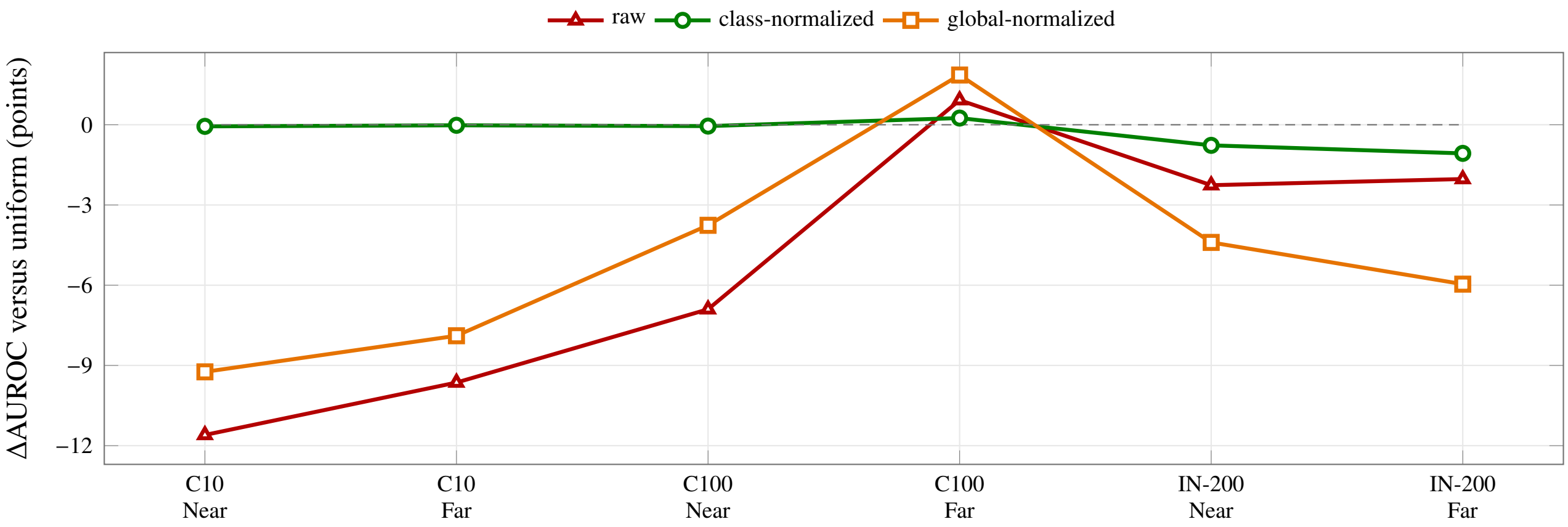
\begin{figure}[H]
\centering
\begin{tikzpicture}
\begin{axis}[
  width=.94\linewidth,height=6.0cm,
  symbolic x coords={C10-N,C10-F,C100-N,C100-F,IN200-N,IN200-F},xtick=data,
  xticklabels={C10\\Near,C10\\Far,C100\\Near,C100\\Far,IN-200\\Near,IN-200\\Far},
  x tick label style={font=\scriptsize,align=center},y tick label style={font=\scriptsize},
  ylabel={$\Delta$AUROC versus uniform (points)},ylabel style={font=\small},
  ymin=-12.7,ymax=2.7,ytick={-12,-9,-6,-3,0},
  axis line style={black!55},grid=major,major grid style={black!9},
  legend style={font=\scriptsize,draw=none,fill=none,legend columns=3,
    at={(.5,1.02)},anchor=south},enlarge x limits=.08]
\addplot[red!70!black,very thick,mark=triangle*,mark options={fill=white}] coordinates
  {(C10-N,-11.60) (C10-F,-9.64) (C100-N,-6.90) (C100-F,.92) (IN200-N,-2.26) (IN200-F,-2.03)};
\addplot[green!50!black,very thick,mark=*,mark options={fill=white}] coordinates
  {(C10-N,-.06) (C10-F,-.02) (C100-N,-.05) (C100-F,.25) (IN200-N,-.77) (IN200-F,-1.07)};
\addplot[orange!90!black,very thick,mark=square*,mark options={fill=white}] coordinates
  {(C10-N,-9.24) (C10-F,-7.89) (C100-N,-3.76) (C100-F,1.86) (IN200-N,-4.40) (IN200-F,-5.96)};
\addplot[black!55,densely dashed,forget plot] coordinates {(C10-N,0) (IN200-F,0)};
\legend{raw,class-normalized,global-normalized}
\end{axis}
\end{tikzpicture}
\caption{Effective-rank anchor weighting relative to the uniform static field.
Raw and global normalization usually reduce AUROC; class-wise normalization
mostly returns to the uniform baseline. The isolated CIFAR-100 Far-OOD gain
does not transfer to its Near-OOD counterpart or ImageNet-200.}
\label{fig:source_mass_failure_v2}
\end{figure}

Table~\ref{tab:mass_full} reports the fitted anchor-weighting variants, and
Fig.~\ref{fig:source_mass_failure_v2} displays changes from uniform weights.
Image weights are the effective rank of mean-centered luminance arrays,
computed as the exponential entropy of normalized singular values.
Bandwidths are refitted for each variant; global weight normalization alone
would leave a fixed static field's ranking unchanged.

\begin{table}[H]
\centering
\caption{Static Gaussian anchor-weighting ablation. Each Near/Far entry is
AUROC/FPR@95 at $T=0$ over three seeds. ``Class'' and ``global'' divide raw
effective-rank weights by their class-wise or global mean.}
\label{tab:mass_full}
\small
\setlength{\tabcolsep}{5pt}
\begin{tabular}{llcc}
\toprule
ID & Anchor weighting & Near-OOD & Far-OOD \\
\midrule
CIFAR-10 & Uniform & 88.59/42.36 & 91.07/31.09 \\
 & Effective rank/raw & 76.99/77.43 & 81.43/55.91 \\
 & Effective rank/class & 88.53/42.85 & 91.04/31.12 \\
 & Effective rank/global & 79.35/71.87 & 83.17/55.49 \\
\midrule
CIFAR-100 & Uniform & 79.46/63.16 & 83.41/51.10 \\
 & Effective rank/raw & 72.55/74.41 & 84.33/50.11 \\
 & Effective rank/class & 79.41/63.35 & 83.66/50.63 \\
 & Effective rank/global & 75.69/71.72 & 85.27/48.29 \\
\midrule
IN-200 & Uniform & 79.44/64.42 & 91.18/31.66 \\
 & Effective rank/raw & 77.17/69.34 & 89.14/41.41 \\
 & Effective rank/class & 78.67/65.05 & 90.11/33.86 \\
 & Effective rank/global & 75.03/69.75 & 85.22/48.57 \\
\bottomrule
\end{tabular}
\end{table}

\subsection{Validation analysis of the static reduction}
\label{app:static_reduction_validation}

The staged analysis uses 5000 ImageNet-1K ID and 1763 OpenImage-O validation
examples. Table~\ref{tab:reduction_bridge_results} reports stage performance;
the following diagnostics quantify adjacent transitions and approximation error.

\begin{table}[H]
\centering
\caption{Full-split transition deltas in the static-reduction sequence. Each
entry is the mean change over the three detector seeds and is written as
AUROC/FPR@95.}
\label{tab:reduction_bridge_transitions}
\small
\setlength{\tabcolsep}{5pt}
\begin{tabular}{lrr}
\toprule
Transition & $\Delta$AUROC & $\Delta$FPR@95 \\
\midrule
S1 $\rightarrow$ S2 & $+13.91$ & $-25.90$ \\
S2 $\rightarrow$ S3 & $+3.18$ & $-7.83$ \\
S3 $\rightarrow$ S4 & $0.00$ & $0.00$ \\
S4 $\rightarrow$ S5 & $+0.95$ & $-3.25$ \\
S5 $\rightarrow$ S6 & $+0.83$ & $-3.78$ \\
MSP $\rightarrow$ S6 & $+7.26$ & $-17.68$ \\
\bottomrule
\end{tabular}
\end{table}

The radial-centroid bound audit evaluates S2 to S3. All six seed/population
combinations satisfy the bound, with
zero violations and negative maximum bound excess. The empirical class-level
MAE ranges from $0.0340$ to $0.0429$, and maximum observed errors range from
$0.209$ to $0.295$. Mean class bounds are approximately $0.685$--$0.692$,
with maxima $0.974$--$1.042$. The bound identifies the dependence on class
dispersion and directional coherence, but is conservative in this experiment.
The uniform Gaussian scores lie in $[0,1]$, whereas the global
$2\epsilon$ from Eq.~\eqref{eq:centroid_margin_preservation} exceeds the
largest possible score margin. Consequently, the
global error bound does not certify any pairwise ranking here. The observed
ranking improvement is an
empirical result, not a consequence of a tight error guarantee.
\FloatBarrier

\subsection{Complete local baseline matrix}
\label{app:complete_empirical_controls}

Tables~\ref{tab:baseline_near_full} and~\ref{tab:baseline_far_full} expose the
complete local baseline matrix used to locate the static controls. Each entry
is mean AUROC/FPR@95 under the same
OpenOOD loaders and checkpoints. ImageNet-1K KNN and ViM were not evaluated
and are marked with a dash. The best AUROC and FPR@95 in each
task column are boldfaced independently.

\begin{table}[H]
\centering
\caption{Complete Near-OOD local comparison including CTM, CADRef, and
LogitGap. Each entry is AUROC/FPR@95 under the unified OpenOOD v1.5 protocol.
FPR@95 is the ID rejection rate at $95\%$ OOD recall (OOD positive).}
\label{tab:baseline_near_full}
\scriptsize
\setlength{\tabcolsep}{4.0pt}
\begin{tabular}{lcccc}
\toprule
Method & CIFAR-10 & CIFAR-100 & IN-200 & IN-1K \\
\midrule
MSP  & $88.03/48.17$ & $80.27/54.80$ & $83.34/54.84$ & $76.02/65.67$ \\
Energy  & $87.58/61.33$ & $80.91/55.60$ & $82.50/60.22$ & $75.89/68.57$ \\
MaxLogit  & $87.52/61.32$ & $81.05/55.46$ & $82.90/59.75$ & $76.46/67.85$ \\
GEN  & $88.20/53.67$ & $81.31/54.41$ & $83.71/55.11$ & $76.85/65.30$ \\
ReAct  & $87.11/63.55$ & $80.77/56.39$ & $81.87/62.46$ & $77.38/66.74$ \\
Scale  & $82.55/80.46$ & $80.99/55.68$ & $84.84/57.30$ & $\mathbf{81.36}/\mathbf{59.77}$ \\
KNN  & $\mathbf{90.64}/\mathbf{34.00}$ & $80.18/61.23$ & $81.57/60.18$ & -- \\
ViM  & $88.68/44.85$ & $74.98/62.63$ & $78.68/59.19$ & -- \\
ASH  & $75.27/86.79$ & $78.19/65.72$ & $82.38/64.88$ & $78.17/63.33$ \\
DICE  & $78.40/70.12$ & $79.38/57.95$ & $81.78/61.87$ & $73.07/72.45$ \\
SHE  & $81.54/79.64$ & $78.95/59.06$ & $80.18/66.79$ & $73.78/73.03$ \\
RMDS  & $86.20/47.20$ & $80.15/55.47$ & $82.57/54.02$ & $76.54/65.24$ \\
RankFeat  & $79.46/60.89$ & $61.88/80.58$ & $56.92/92.04$ & $50.99/91.83$ \\
CADRef  & $88.97/39.46$ & $79.63/57.18$ & $\mathbf{85.57}/\mathbf{52.97}$ & $79.62/61.57$ \\
LogitGap  & $88.72/52.34$ & $\mathbf{81.37}/\mathbf{54.36}$ & $83.37/57.63$ & $76.70/66.47$ \\
Centroid-only & $89.35/38.16$ & $80.12/59.07$ & $82.57/60.11$ & $74.34/69.41$ \\
\method\ & $89.27/38.15$ & $80.35/58.75$ & $83.33/58.62$ & $76.29/67.93$ \\
CTM  & $89.49/37.29$ & $80.16/59.18$ & $83.13/58.21$ & $75.69/67.86$ \\
\bottomrule
\end{tabular}
\end{table}

\begin{table}[H]
\centering
\caption{Complete Far-OOD local comparison including CTM, CADRef, and LogitGap.
Each entry is AUROC/FPR@95 under the unified OpenOOD v1.5 protocol.
FPR@95 is the ID rejection rate at $95\%$ OOD recall (OOD positive).}
\label{tab:baseline_far_full}
\scriptsize
\setlength{\tabcolsep}{4.0pt}
\begin{tabular}{lcccc}
\toprule
Method & CIFAR-10 & CIFAR-100 & IN-200 & IN-1K \\
\midrule
MSP  & $90.73/31.72$ & $77.76/58.71$ & $90.13/35.44$ & $85.23/51.46$ \\
Energy  & $91.21/41.70$ & $79.77/56.58$ & $90.86/34.85$ & $89.47/38.38$ \\
MaxLogit  & $91.10/41.67$ & $79.67/56.73$ & $91.11/34.03$ & $89.58/38.21$ \\
GEN  & $91.35/34.73$ & $79.68/56.71$ & $91.36/32.11$ & $89.77/35.62$ \\
ReAct  & $90.42/44.90$ & $80.39/54.21$ & $92.31/28.50$ & $93.67/26.31$ \\
Scale  & $86.39/67.54$ & $81.42/54.08$ & $93.98/26.47$ & $\mathbf{96.53}/\mathbf{16.53}$ \\
KNN  & $92.96/\mathbf{24.28}$ & $82.40/53.64$ & $93.16/27.27$ & -- \\
ViM  & $\mathbf{93.48}/25.06$ & $81.70/50.73$ & $91.26/27.20$ & -- \\
ASH  & $78.49/79.03$ & $80.58/59.20$ & $93.90/27.30$ & $95.74/19.53$ \\
DICE  & $84.29/51.71$ & $80.01/56.25$ & $90.80/36.51$ & $90.95/41.82$ \\
SHE  & $85.32/66.48$ & $76.92/64.12$ & $89.81/42.16$ & $90.92/41.43$ \\
RMDS  & $91.30/29.41$ & $82.92/52.82$ & $88.06/32.45$ & $86.95/39.65$ \\
RankFeat  & $75.87/57.44$ & $67.10/69.45$ & $38.22/97.72$ & $53.93/87.15$ \\
CADRef  & $91.51/28.19$ & $\mathbf{83.49}/\mathbf{50.33}$ & $\mathbf{95.50}/\mathbf{19.58}$ & $96.27/16.85$ \\
LogitGap  & $91.88/33.41$ & $79.74/55.95$ & $91.24/32.17$ & $89.69/36.02$ \\
Centroid-only & $91.62/28.24$ & $82.52/52.43$ & $94.18/23.72$ & $94.50/24.54$ \\
\method\ & $91.58/28.18$ & $81.70/52.88$ & $94.02/24.17$ & $94.85/23.60$ \\
CTM  & $91.68/27.76$ & $82.62/52.50$ & $93.89/24.52$ & $95.20/21.68$ \\
\bottomrule
\end{tabular}
\end{table}

\begin{table}[!htbp]
\centering
\begin{minipage}[t]{.57\textwidth}
\centering
\caption{Recent and centroid references on ImageNet-1K/DenseNet-121. CADRef,
LogitGap, and CTM are single fixed-model estimates; centroid-only and fixed
fusion show mean$\pm$standard deviation over three sampled setup banks on the
same model. Boldface compares the displayed methods. FPR@95 uses OOD as
the positive class, as defined in Section~\ref{sec:Experimental Protocol}.}
\label{tab:ctm_densenet_reference}
\scriptsize
\setlength{\tabcolsep}{3.5pt}
\begin{tabular}{llcc}
\toprule
Group & Method & AUROC & FPR@95 \\
\midrule
Near & CADRef  & $\mathbf{76.13}$ & $\mathbf{66.60}$ \\
     & LogitGap  & 75.49 & 67.42 \\
     & CTM  & 72.91 & 71.61 \\
     & Centroid-only & $71.18\pm0.35$ & $73.29\pm0.56$ \\
     & \method\ & $73.88\pm0.31$ & $70.80\pm0.88$ \\
Far  & CADRef  & $\mathbf{95.32}$ & $\mathbf{21.46}$ \\
     & LogitGap  & 89.14 & 39.69 \\
     & CTM  & 94.02 & 25.88 \\
     & Centroid-only & $93.01\pm0.53$ & $29.59\pm1.63$ \\
     & \method\ & $93.95\pm0.43$ & $26.97\pm1.46$ \\
\bottomrule
\end{tabular}
\end{minipage}\hfill
\begin{minipage}[t]{.40\textwidth}
\centering
\caption{Additional CTM precision--recall metrics on ImageNet-1K/DenseNet-121.
AUROC and FPR@95 are reported in Table~\ref{tab:per_dataset_imagenet}.
The ID accuracy is $74.86\%$.}
\label{tab:ctm_densenet_datasets}
\small
\begin{tabular}{lrr}
\toprule
OOD dataset & AUPR-IN & AUPR-OUT \\
\midrule
SSB-hard & 63.73 & 64.98 \\
NINCO & 96.83 & 36.02 \\
iNaturalist & 98.65 & 82.65 \\
Textures & 99.67 & 87.78 \\
OpenImage-O & 95.99 & 76.59 \\
\bottomrule
\end{tabular}
\end{minipage}
\end{table}

\subsection{Complete-coverage ranking}

For each ID/group pair, all available methods are ranked by AUROC and FPR@95,
including KNN and ViM on the six tasks they cover. Ties receive the minimum
rank. Table~\ref{tab:ranking_full} reports mean ranks for the 16 scoring
configurations with complete eight-task coverage, including centroid-only.
Ranks use the two-decimal means displayed in the complete baseline matrices.
The composite averages the AUROC and FPR@95 ranks.

\begin{table}[H]
\centering
\caption{Mean ranks for scoring configurations with complete eight-task
coverage, ordered by AUROC rank. Per-task rankings include all available
configurations, including KNN and ViM where evaluated; only complete-coverage
configurations are listed. Ranks use the two-decimal means displayed in
Tables~\ref{tab:baseline_near_full} and~\ref{tab:baseline_far_full}, with
minimum ranks for ties. Lower is better.}
\label{tab:ranking_full}
\small
\setlength{\tabcolsep}{3pt}
\begin{tabular}{lrrr}
\toprule
Configuration & AUROC rank & FPR rank & Composite \\
\midrule
CADRef & \textbf{4.000} & \textbf{3.250} & \textbf{3.625} \\
\method & 5.750 & 6.375 & 6.063 \\
CTM & 6.000 & 6.000 & 6.000 \\
Scale & 6.000 & 7.625 & 6.813 \\
Centroid-only & 6.750 & 7.250 & 7.000 \\
LogitGap & 6.875 & 7.750 & 7.313 \\
GEN & 7.375 & 7.750 & 7.563 \\
ReAct & 9.125 & 10.375 & 9.750 \\
MaxLogit & 10.000 & 10.375 & 10.188 \\
RMDS & 10.250 & 6.750 & 8.500 \\
ASH & 10.375 & 12.375 & 11.375 \\
Energy & 10.625 & 11.625 & 11.125 \\
MSP & 11.250 & 9.125 & 10.188 \\
DICE & 13.625 & 13.500 & 13.563 \\
SHE & 14.625 & 15.375 & 15.000 \\
RankFeat & 17.250 & 16.250 & 16.750 \\
\bottomrule
\end{tabular}
\end{table}

\subsection{Per-dataset OOD detection results}

Tables~\ref{tab:per_dataset_cifar} and~\ref{tab:per_dataset_imagenet}
report the named OOD datasets underlying the Near-/Far-OOD aggregates in the
main text. Each cell is AUROC/FPR@95. CIFAR-10, CIFAR-100, and ImageNet-200
values are mean$\pm$standard deviation over three official model seeds.
ImageNet-1K uses a fixed pretrained backbone: MSP and Scale are therefore
single runs, while the centroid--MSP baseline reports variation over three
independently sampled centroid banks. Within each dataset row, the
highest AUROC and the lowest FPR@95 are boldfaced independently.

\begin{figure}[H]
\centering
\begin{tikzpicture}
\begin{groupplot}[
  group style={group size=2 by 1,horizontal sep=1.05cm},
  width=.47\textwidth,height=5.4cm,xbar,
  symbolic y coords={peer,Tiny,MNIST,SVHN,Places,Textures},ytick=data,
  yticklabels={peer CIFAR,Tiny ImageNet,MNIST,SVHN,Places365,Textures},
  y tick label style={font=\scriptsize},x tick label style={font=\scriptsize},
  title style={font=\small\bfseries},label style={font=\small},
  xlabel={$\Delta$AUROC versus MSP},xmin=-1.3,xmax=6.2,
  xtick={-1,0,1,2,3,4,5,6},axis line style={black!55},
  grid=major,major grid style={black!9},enlarge y limits=.10,
  nodes near coords,nodes near coords style={font=\tiny},
  point meta=explicit symbolic]
\nextgroupplot[title={(a) CIFAR-10 ID}]
\addplot[fill=blue!62,draw=blue!70!black] coordinates
  {(1.26,peer)[1.26] (1.23,Tiny)[1.23] (.21,MNIST)[.21]
   (.11,SVHN)[.11] (1.23,Places)[1.23] (1.87,Textures)[1.87]};
\addplot[black!55,densely dashed,forget plot] coordinates {(0,peer) (0,Textures)};

\nextgroupplot[title={(b) CIFAR-100 ID},yticklabels={}]
\addplot[fill=blue!62,draw=blue!70!black] coordinates
  {(1.02,Tiny)[1.02] (4.33,MNIST)[4.33] (4.97,SVHN)[4.97]
   (.88,Places)[.88] (5.58,Textures)[5.58]};
\addplot[fill=red!62,draw=red!70!black] coordinates {(-.86,peer)[-.86]};
\addplot[black!55,densely dashed,forget plot] coordinates {(0,peer) (0,Textures)};
\end{groupplot}
\end{tikzpicture}
\caption{Per-dataset AUROC change of the centroid--MSP baseline relative to MSP on
CIFAR. Blue bars indicate improvement and red indicates degradation. Gains are
broad but heterogeneous; peer CIFAR under CIFAR-100 is the only negative case.
Exact AUROC/FPR@95 values are reported in Table~\ref{tab:per_dataset_cifar}.}
\label{fig:per_dataset_cifar_gain_v2}
\end{figure}

\begin{table}[H]
\centering
\caption{Per-dataset CIFAR results under the unified OpenOOD v1.5 protocol.
Each entry is AUROC/FPR@95; all values are mean$\pm$standard deviation over
three official ResNet-18 seeds.}
\label{tab:per_dataset_cifar}
\scriptsize
\setlength{\tabcolsep}{2.4pt}
\begin{tabular}{lllcccc}
\toprule
ID & OOD dataset & Type & MSP & Scale & CTM & \method \\
\midrule
CIFAR-10 & CIFAR-100     & Near & $87.19\pm.40/53.09\pm6.02$ & $81.27\pm.91/81.78\pm4.87$ & $\mathbf{88.66\pm.17}/\mathbf{40.79\pm.91}$ & $88.45\pm.18/41.96\pm1.41$ \\
         & Tiny ImageNet & Near & $88.87\pm.24/43.25\pm3.64$ & $83.84\pm.02/79.14\pm4.95$ & $\mathbf{90.32\pm.30}/\mathbf{33.79\pm1.09}$ & $90.10\pm.33/34.33\pm1.43$ \\
         & MNIST         & Far  & $92.63\pm1.92/\mathbf{23.64\pm7.13}$ & $90.58\pm3.62/48.69\pm18.84$ & $92.82\pm.49/23.84\pm2.14$ & $\mathbf{92.84\pm.45}/23.72\pm1.75$ \\
         & SVHN          & Far  & $91.46\pm.50/\mathbf{25.80\pm2.01}$ & $84.63\pm3.74/70.54\pm10.67$ & $91.45\pm.28/26.72\pm1.17$ & $\mathbf{91.57\pm.60}/27.05\pm1.79$ \\
         & Places365     & Far  & $88.92\pm.57/42.47\pm4.68$ & $86.41\pm2.35/70.51\pm7.32$ & $\mathbf{90.52\pm.30}/\mathbf{33.46\pm.70}$ & $90.15\pm.32/34.77\pm.65$ \\
         & Textures      & Far  & $89.89\pm.87/34.97\pm5.71$ & $83.93\pm.42/80.40\pm4.80$ & $\mathbf{91.93\pm.26}/\mathbf{27.03\pm.33}$ & $91.76\pm.28/27.16\pm.63$ \\
\midrule
CIFAR-100& CIFAR-10      & Near & $78.47\pm.08/\mathbf{58.90\pm1.14}$ & $\mathbf{79.26\pm.12}/59.11\pm1.03$ & $77.25\pm.21/68.36\pm.35$ & $77.61\pm.14/67.82\pm.71$ \\
         & Tiny ImageNet & Near & $82.07\pm.20/50.69\pm.43$ & $82.71\pm.18/52.25\pm.89$ & $83.08\pm.25/50.00\pm.29$ & $\mathbf{83.09\pm.23}/\mathbf{49.68\pm.16}$ \\
         & MNIST         & Far  & $76.08\pm2.28/57.24\pm5.72$ & $80.27\pm1.42/51.64\pm4.37$ & $\mathbf{81.32\pm2.20}/\mathbf{50.19\pm5.15}$ & $80.41\pm2.07/50.47\pm4.80$ \\
         & SVHN          & Far  & $78.42\pm1.09/59.08\pm3.10$ & $84.45\pm1.33/\mathbf{49.27\pm3.18}$ & $\mathbf{84.56\pm1.59}/49.97\pm4.04$ & $83.39\pm.85/50.43\pm2.68$ \\
         & Places365     & Far  & $79.23\pm.35/\mathbf{56.62\pm1.06}$ & $\mathbf{80.47\pm.28}/56.98\pm.83$ & $80.43\pm.50/58.80\pm1.04$ & $80.11\pm.37/59.00\pm1.13$ \\
         & Textures      & Far  & $77.32\pm.86/61.90\pm1.59$ & $80.50\pm.86/58.44\pm2.41$ & $\mathbf{84.17\pm.66}/\mathbf{51.05\pm2.28}$ & $82.90\pm.65/51.63\pm1.91$ \\
\bottomrule
\end{tabular}
\end{table}

\begin{figure}[H]
\centering
\begin{tikzpicture}
\begin{groupplot}[
  group style={group size=2 by 1,horizontal sep=1.0cm},
  width=.47\textwidth,height=6.0cm,xbar,
  symbolic y coords={SSB,NINCO,iNat,Textures,OpenImage},ytick=data,
  yticklabels={SSB-hard,NINCO,iNaturalist,Textures,OpenImage-O},
  y tick label style={font=\scriptsize},x tick label style={font=\scriptsize},
  title style={font=\small\bfseries},label style={font=\small},
  axis line style={black!55},grid=major,major grid style={black!9},
  enlarge y limits=.12,
  legend style={font=\scriptsize,draw=none,fill=none,legend columns=3,
    at={(.5,1.18)},anchor=south}]
\nextgroupplot[title={(a) AUROC change},xlabel={$\Delta$AUROC versus MSP},
  xmin=-6,xmax=17.5,xtick={-5,0,5,10,15}]
\addplot[fill=blue!35,draw=blue!55!black,bar shift=-3pt] coordinates
  {(-.79,SSB) (.78,NINCO) (3.32,iNat) (5.67,Textures) (2.67,OpenImage)};
\addplot[fill=blue!65,draw=blue!75!black,bar shift=0pt] coordinates
  {(-2.94,SSB) (3.49,NINCO) (6.94,iNat) (14.82,Textures) (7.08,OpenImage)};
\addplot[fill=orange!70,draw=orange!85!black,bar shift=3pt] coordinates
  {(-5.19,SSB) (2.31,NINCO) (5.90,iNat) (15.97,Textures) (6.20,OpenImage)};
\addplot[black!55,densely dashed,forget plot] coordinates {(0,SSB) (0,OpenImage)};
\legend{IN-200/R18,IN-1K/R50,IN-1K/D121}

\nextgroupplot[title={(b) FPR@95 reduction},xlabel={MSP minus centroid--MSP FPR@95},
  yticklabels={},xmin=-9,xmax=52,xtick={0,10,20,30,40,50}]
\addplot[fill=blue!35,draw=blue!55!black,bar shift=-3pt] coordinates
  {(-6.29,SSB) (-1.28,NINCO) (8.93,iNat) (18.45,Textures) (6.41,OpenImage)};
\addplot[fill=blue!65,draw=blue!75!black,bar shift=0pt] coordinates
  {(-6.29,SSB) (1.77,NINCO) (21.97,iNat) (45.07,Textures) (16.56,OpenImage)};
\addplot[fill=orange!70,draw=orange!85!black,bar shift=3pt] coordinates
  {(-7.57,SSB) (.07,NINCO) (17.61,iNat) (48.91,Textures) (12.24,OpenImage)};
\addplot[black!55,densely dashed,forget plot] coordinates {(0,SSB) (0,OpenImage)};
\end{groupplot}
\end{tikzpicture}
\caption{Per-dataset change of the centroid--MSP baseline relative to MSP on the
ImageNet benchmarks. Positive values are favorable in both panels. Grouping
the three backbone/scale settings within each dataset exposes the shared
pattern: large Far-OOD gains, especially on Textures, and a consistent loss on
SSB-hard. Exact values appear in Table~\ref{tab:per_dataset_imagenet}.}
\label{fig:per_dataset_imagenet_gain_v2}
\end{figure}
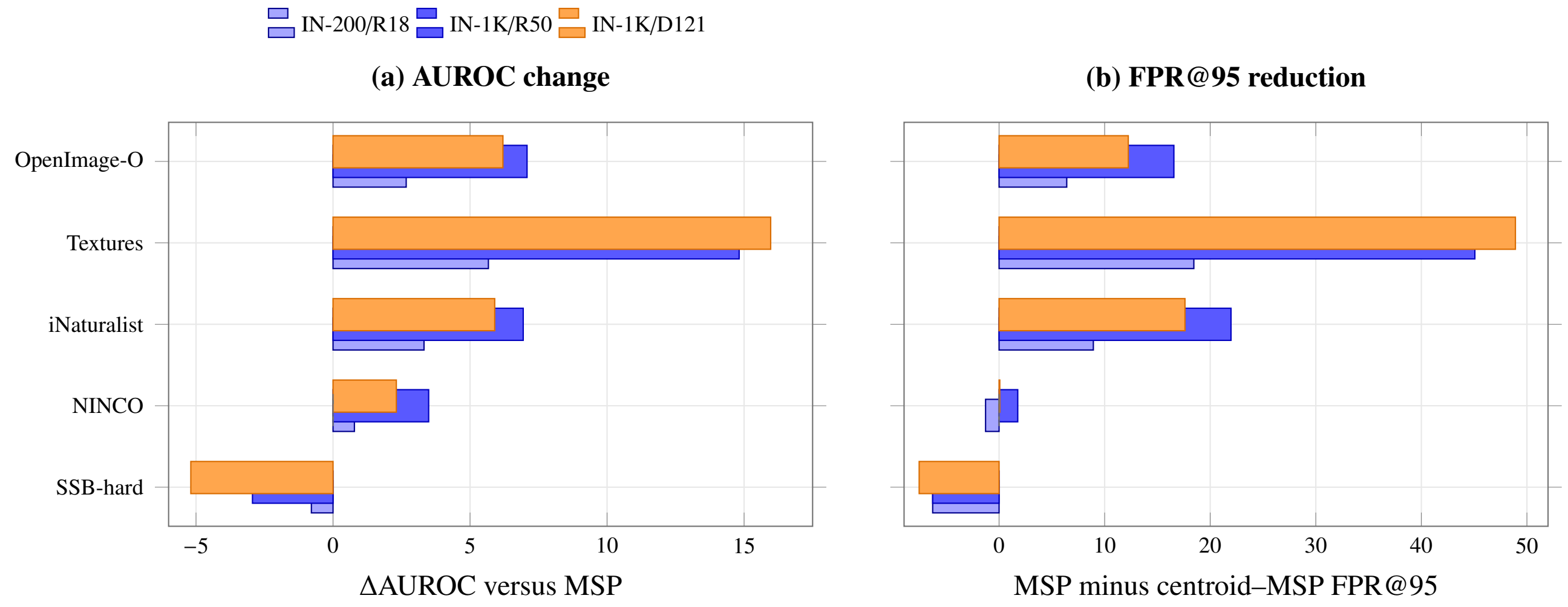

\begin{table}[H]
\centering
\caption{Per-dataset ImageNet results. Each entry is AUROC/FPR@95. ImageNet-200
uses three official ResNet-18 seeds. ImageNet-1K MSP, Scale, and CTM use one
fixed backbone, whereas \method\ varies three centroid banks.}
\label{tab:per_dataset_imagenet}
\scriptsize
\setlength{\tabcolsep}{2.4pt}
\begin{tabular}{lllcccc}
\toprule
ID/backbone & OOD dataset & Type & MSP & Scale & CTM & \method \\
\midrule
IN-200/R18 & SSB-hard     & Near & $80.38\pm.04/\mathbf{66.00\pm.14}$ & $\mathbf{82.08\pm.33}/67.39\pm.78$ & $79.43\pm.25/71.32\pm.29$ & $79.59\pm.24/72.29\pm.36$ \\
           & NINCO        & Near & $86.29\pm.13/\mathbf{43.67\pm.93}$ & $\mathbf{87.59\pm.36}/47.21\pm1.55$ & $86.83\pm.17/45.10\pm1.31$ & $87.07\pm.16/44.95\pm1.35$ \\
           & iNaturalist  & Far  & $92.80\pm.30/26.49\pm.91$ & $95.79\pm.56/18.41\pm2.30$ & $95.90\pm.18/18.70\pm.99$ & $\mathbf{96.12\pm.33}/\mathbf{17.56\pm.68}$ \\
           & Textures     & Far  & $88.36\pm.16/44.58\pm.86$ & $94.11\pm.08/29.80\pm.64$ & $\mathbf{94.42\pm.07}/\mathbf{25.13\pm.85}$ & $94.03\pm.20/26.13\pm.87$ \\
           & OpenImage-O  & Far  & $89.24\pm.02/35.24\pm.27$ & $\mathbf{92.04\pm.30}/31.20\pm1.32$ & $91.34\pm.12/29.74\pm.70$ & $91.91\pm.06/\mathbf{28.83\pm.27}$ \\
\midrule
IN-1K/R50  & SSB-hard     & Near & $72.09/74.48$ & $\mathbf{77.35}/\mathbf{67.73}$ & $68.53/80.07$ & $69.15\pm.20/80.77\pm.30$ \\
           & NINCO        & Near & $79.95/56.87$ & $\mathbf{85.37}/\mathbf{51.81}$ & $82.85/55.65$ & $83.44\pm.25/55.10\pm.89$ \\
           & iNaturalist  & Far  & $88.41/43.35$ & $\mathbf{98.02}/\mathbf{9.50}$  & $95.83/19.33$ & $95.35\pm.56/21.38\pm2.36$ \\
           & Textures     & Far  & $82.43/60.90$ & $97.63/\mathbf{11.90}$ & $\mathbf{97.75}/12.71$ & $97.25\pm.09/15.83\pm.78$ \\
           & OpenImage-O  & Far  & $84.86/50.14$ & $\mathbf{93.95}/\mathbf{28.18}$ & $92.01/33.00$ & $91.94\pm.46/33.58\pm1.34$ \\
\midrule
IN-1K/D121 & SSB-hard     & Near & $71.11/75.52$ & $\mathbf{72.75}/\mathbf{75.26}$ & $64.95/82.91$ & $65.92\pm.24/83.09\pm.26$ \\
           & NINCO        & Near & $79.53/58.58$ & $\mathbf{83.56}/\mathbf{55.78}$ & $80.86/60.31$ & $81.84\pm.39/58.51\pm1.51$ \\
           & iNaturalist  & Far  & $88.61/42.78$ & $\mathbf{97.25}/\mathbf{13.94}$ & $94.80/23.09$ & $94.51\pm.77/25.17\pm3.08$ \\
           & Textures     & Far  & $81.28/64.32$ & $96.17/19.47$ & $\mathbf{97.49/13.44}$ & $97.25\pm.12/15.41\pm.93$ \\
           & OpenImage-O  & Far  & $83.88/52.59$ & $\mathbf{92.74}/\mathbf{33.02}$ & $89.78/41.10$ & $90.08\pm.58/40.35\pm1.73$ \\
\bottomrule
\end{tabular}
\end{table}

\subsection{Validation mechanism, sample efficiency, and calibration budget}

The following tables and plots give the numerical detail for
Section~\ref{sec:Controlled Mechanism and Robustness Studies}. Pointwise
best weights maximize mean AUROC on the designated validation holdout
retrospectively; the evaluated rule remains fixed at $\alpha=0.8$.

\begin{table}[H]
\centering
\caption{Exact values for the controlled nested class-count validation study.
$\alpha^*$ maximizes mean holdout AUROC at each class count. The two $\Delta$
columns evaluate the single pre-test choice $\alpha=0.8$ against MSP
($\alpha=0$) and centroid-only ($\alpha=1$); positive values favor the fixed
mixture. Results use one nested class sequence.}
\label{tab:class_scale}
\scriptsize
\setlength{\tabcolsep}{3.5pt}
\begin{tabular}{rrrrc}
\toprule
Classes & Pointwise $\alpha^*$ & $\Delta$ vs. MSP & $\Delta$ vs. centroid & $0.8=\alpha^*$ \\
\midrule
10   & 1.0 & $+0.64$ & $-1.00$ & -- \\
25   & 0.9 & $+2.39$ & $-0.14$ & -- \\
50   & \textbf{0.8} & $+2.37$ & $+0.33$ & \checkmark \\
100  & \textbf{0.8} & $+2.76$ & $+0.67$ & \checkmark \\
200  & \textbf{0.8} & $+4.74$ & $+0.45$ & \checkmark \\
500  & \textbf{0.8} & $+5.45$ & $+0.69$ & \checkmark \\
1000 & 0.7 & $+6.45$ & $+0.86$ & -- \\
\bottomrule
\end{tabular}
\end{table}

\begin{figure}[H]
\centering
\begin{tikzpicture}
\begin{groupplot}[
  group style={group size=2 by 1,horizontal sep=1.05cm},
  width=.47\textwidth,height=4.9cm,xtick={1,2,4,6,8,12},
  xlabel={Centroid samples per class},
  x tick label style={font=\scriptsize},y tick label style={font=\scriptsize},
  title style={font=\small\bfseries},label style={font=\small},
  axis line style={black!55},grid=major,major grid style={black!9},
  legend style={font=\scriptsize,draw=none,fill=none,legend columns=2,
    at={(.5,1.18)},anchor=south}]
\nextgroupplot[title={(a) Validation AUROC},ylabel={AUROC (\%)},
  ymin=86,ymax=92.2,ytick={86,88,90,92}]
\addplot[blue!70!black,very thick,mark=*,mark options={fill=white}]
  coordinates {(1,89.04) (2,90.10) (4,90.97) (6,91.50) (8,91.65) (12,91.78)};
\addplot[orange!90!black,very thick,mark=square*,mark options={fill=white}]
  coordinates {(1,86.74) (2,88.92) (4,90.60) (6,91.32) (8,91.53) (12,91.72)};
\legend{holdout envelope,fixed $\alpha=0.8$}

\nextgroupplot[title={(b) Validation FPR@95},ylabel={FPR@95 (\%)},
  ymin=30,ymax=52,ytick={30,35,40,45,50}]
\addplot[blue!70!black,very thick,mark=*,mark options={fill=white}]
  coordinates {(1,45.64) (2,40.16) (4,37.69) (6,34.66) (8,33.81) (12,32.37)};
\addplot[orange!90!black,very thick,mark=square*,mark options={fill=white}]
  coordinates {(1,50.33) (2,42.47) (4,39.79) (6,34.90) (8,34.52) (12,32.91)};
\end{groupplot}
\end{tikzpicture}
\caption{ImageNet-1K validation-only sample efficiency over three centroid
seeds. Six samples per class capture most of the 12-sample performance, and
the fixed $\alpha=0.8$ curve nearly meets the retrospective holdout envelope
from six samples onward. The blue curve maximizes mean holdout AUROC over
$\alpha$ separately at each bank size; panel (b) uses the corresponding weights.}
\label{fig:sample_efficiency_v2}
\end{figure}
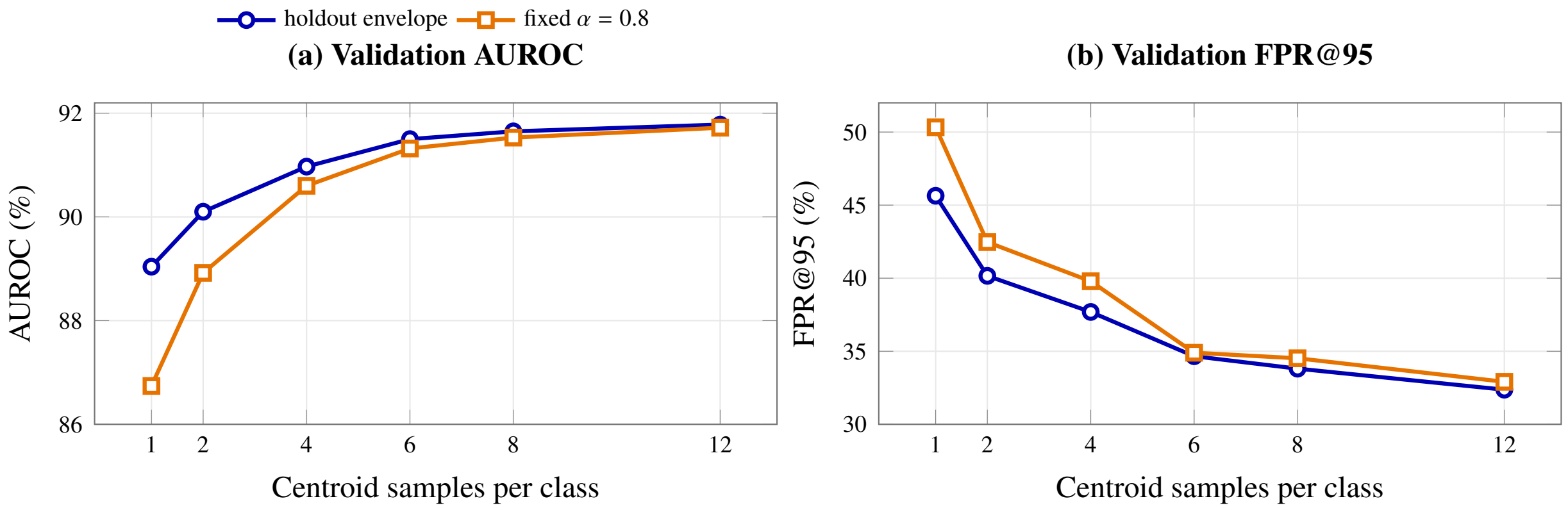

\begin{table}[H]
\centering
\caption{ImageNet-1K validation-only calibration-budget sensitivity of the
centroid--MSP baseline. Columns vary the number of ID calibration images per
class; each entry is AUROC/FPR@95. One ID image per class is held out for
evaluation, calibration subsets are nested within the other four, and all
OOD validation examples are evaluated. The centroid bank is fixed. Positive
affine standardization leaves either component's rankings unchanged.}
\label{tab:calibration_budget}
\small
\setlength{\tabcolsep}{6pt}
\begin{tabular}{lcccc}
\toprule
Method & 1 image & 2 images & 3 images & 4 images \\
\midrule
Fixed $\alpha=0.8$ & 92.185/31.17 & 92.184/31.13 & 92.185/31.13 & 92.183/31.13 \\
\bottomrule
\end{tabular}
\end{table}

\begin{table}[H]
\centering
\caption{Validation-holdout mechanism audit on the five natural
dataset/backbone targets. ``Corr.'' is Pearson correlation between MSP and
centroid scores on the designated OOD validation holdout. Score cells are
AUROC/FPR@95; values aggregate three centroid seeds where applicable.}
\label{tab:natural_validation_mechanism}
\scriptsize
\setlength{\tabcolsep}{3.4pt}
\begin{tabular}{lrrrrrr}
\toprule
Target & Classes & OOD corr. & Best $\alpha$ & MSP & Fixed $\alpha=0.8$ & Centroid-only \\
\midrule
CIFAR-10/R18 & 10 & $.880\pm.002$ & 1.0 & 88.65/42.73 & 89.72/36.40 & 89.80/36.53 \\
CIFAR-100/R18 & 100 & $.828\pm.003$ & 1.0 & 83.62/46.53 & 84.99/45.07 & 85.00/44.67 \\
IN-200/R18 & 200 & $.677\pm.016$ & 1.0 & 87.81/38.33 & 90.75/30.94 & 90.77/31.44 \\
IN-1K/R50 & 1000 & $.410\pm.015$ & .8 & 85.27/47.13 & \textbf{91.72/32.91} & 90.86/36.89 \\
IN-1K/D121 & 1000 & $.315\pm.003$ & .6 & 83.77/51.90 & \textbf{90.31/39.16} & 88.81/42.57 \\
\bottomrule
\end{tabular}
\end{table}

Recomputing the 12-sample centroids reproduces the archived checkpoint within
$4.47\times10^{-8}$ maximum absolute error for each of the three seeds.

\subsection{Extended batch-one efficiency comparison}

Table~\ref{tab:efficiency_extended} reports the complete batch-one latency
ratios underlying the abbreviated discussion in Section~\ref{sec:Experiments}.
All measurements use the same RTX 4090 D, include the backbone forward pass,
exclude data loading, and use MSP measured in the corresponding timing session
as ratio one. CTM and
the centroid--MSP baseline are measured in one paired runner on identical preloaded
inputs.

\begin{table}[!htbp]
\centering
\caption{End-to-end batch-one latency relative to MSP. Lower is better.
Per-image latency measures online inference; setup-sample counts for CTM and
the centroid--MSP baseline are described in the accompanying resource audit.}
\label{tab:efficiency_extended}
\small
\setlength{\tabcolsep}{6pt}
\begin{tabular}{lcccc}
\toprule
Method & CIFAR-10 & CIFAR-100 & IN-200 & IN-1K \\
\midrule
CTM & \textbf{0.994} & 1.043 & 1.037 & 1.016 \\
MSP & 1.000 & \textbf{1.000} & 1.000 & \textbf{1.000} \\
\method & 1.041 & 1.085 & 1.090 & 1.039 \\
SHE & 1.045 & 1.180 & \textbf{0.776} & 1.069 \\
RMDS & 1.137 & 1.146 & 1.103 & 1.509 \\
Scale & 1.141 & 1.122 & 1.158 & 1.516 \\
DICE & 1.146 & 1.120 & 1.138 & 1.670 \\
ASH & 1.149 & 1.115 & 1.207 & 1.735 \\
RankFeat & 3.073 & 3.045 & 4.741 & 4.505 \\
\bottomrule
\end{tabular}
\end{table}

\subsection{Implementation details}
\label{app:implementation_details}

For the small-bank static control, normalized feature directions are averaged
within each class and renormalized to obtain $\mu_c$. The query statistic is
$G(x)=\max_cq(x)^\top\mu_c$. If $m(x)$ is MSP and $\widetilde m,\widetilde G$
denote pooled ID-standardized components, the reported fixed control is
$(1-\alpha)\widetilde m+\alpha\widetilde G$ with $\alpha=0.8$.

\paragraph{Training banks and candidate classes.}
All feature directions are normalized before anchor or centroid construction.
The original trajectory settings below are distinct from the mechanism audits.
For the original CIFAR trajectories, 80 anchors per class are sampled from the extracted training
features using a seeded random permutation. All classes participate in both
bandwidth fitting and inference. For ImageNet, the setup bank contains 12
training examples per class, and five are selected by a seeded random
permutation as field anchors. The normalized mean of all 12 feature directions
defines the class centroid used for candidate selection. At inference, the
20 classes with the largest cosine similarities to the initial query are
retained for both force computation and trajectory scoring. This candidate
set remains fixed as the query moves. The centroid--MSP baseline uses all
class centroids, without candidate truncation.

The score-aligned mechanism audit instead uses five anchors from 12 setup
features per class on every benchmark, fits on that setup bank, and uses
$\min(C,20)$ initial candidates. The stability probe uses normalized means
of all 12-feature setup banks on every benchmark, compares all class centroids,
and does not evaluate an anchor kernel during the probe.

\paragraph{Bandwidth fitting.}
The reported trajectory grids, anchor-weighting study, and five-kernel
comparison fit bandwidths using zero-step affinities. With the backbone,
anchors, and anchor weights fixed, the objective is
\begin{equation}
 \mathcal L(\sigma)=-\frac{1}{N}\sum_{i=1}^{N}
 \log\frac{\exp\{K A_{y_i}(q_i)\}}
 {\sum_{c\in\mathcal J_i^{\rm fit}}\exp\{K A_c(q_i)\}},
 \label{eq:bandwidth_fitting_loss}
\end{equation}
where $(q_i,y_i)$ are ID fitting features and labels.
On CIFAR, $\mathcal J_i^{\rm fit}$ contains every class and fitting uses the
extracted training features. On ImageNet, fitting uses the setup bank and
$\mathcal J_i^{\rm fit}$ contains the true class and the 19 other classes
with highest centroid cosine. The true class is included only because labels
are available during fitting; inference uses the 20 highest-scoring classes
without this intervention.

The factor $K$ matches the implementation's summed-anchor affinities. It does
not change the ordering of a fixed field, but it matters in the fitting loss
and force magnitude. We optimize log-bandwidths with Adam at an initial
learning rate of $10^{-3}$ and cosine learning-rate decay. Bandwidths are
initialized to $1.0$ on CIFAR and $0.5$ on ImageNet and constrained to
$[0.05,4.0]$. CIFAR uses 25 epochs with batches of 64; ImageNet uses 10 epochs
with batches of 32. No trajectory is differentiated through in this fitting
objective, even when inference uses a nonzero step count. The static
centroid--MSP baseline requires neither bandwidth fitting nor integration.

\paragraph{Score calibration and fusion.}
The aligned audit uses a seeded floor-half split within each ID class and a
separate floor-half OOD split for tuning, with the complements held out.
Current-winner $T\in\{1,3,5,10\}$ is selected by mean ImageNet-1K tuning
AUROC over three banks, breaking ties by lower FPR@95 and then smaller $T$.
The selected $T=5$ is transferred without retuning. The no-switch audit uses
the full validation caches, rather than only the evaluation holdouts.
Per seed, the ImageNet-1K selection and holdout counts are respectively
2000/881 and 3000/882 (ID/OOD). Target holdouts contain 500/500 for CIFAR-10,
500/486 for CIFAR-100, and 600/882 for ImageNet-200; no target-side selection
is performed for this study.

The probe uses a separate partition with calibration seed zero: two ID
examples per class for fit, one for selection, and the remainder for the
holdout. A separate OOD permutation has split endpoints
$\lfloor.35N_{\rm out}\rfloor$ and $\lfloor.60N_{\rm out}\rfloor$; OOD-fit
examples are not used for calibration. On ResNet-50 source selection data,
$\eta\in\{0,.01,.02,.04,.08,.16\}$ is chosen by mean AUROC over three banks,
breaking ties by smaller $\eta$. The selected $.04$ is fixed for all other
settings; target selection subsets are unused. Probe statistics and AUROC
changes are measured on the same holdout, so their agreement is
descriptive rather than an independent prediction test.
Per seed, ImageNet-1K uses 2000 ID-fit examples, 1000/440 selection examples,
and 2000/706 holdout examples (ID/OOD), for both backbones; only ResNet-50
is used to select $\eta$. CIFAR-10, CIFAR-100, and ImageNet-200 use
20, 200, and 400 ID-fit examples, respectively, with holdout counts of
970/400, 700/389, and 400/706. ID-fit examples standardize the probe scores;
training-feature banks are separate.

For each benchmark and backbone, the two component scores are standardized
using the same deterministic, class-stratified subset of ID validation data.
The stored statistics are the pooled means and population standard deviations;
inference denominators are floored at $10^{-12}$. Selecting the maximum-cosine
rule and $\alpha=0.8$ uses an ImageNet-1K ID/OOD tuning split for the static
baseline,
with $\alpha$ scanned on
$\{0,.05,\ldots,1\}$ and only the ID tuning part used for standardization. The
rule is checked on the complementary holdout. The probe inherits this fixed
$\alpha$; its separate selection split is used to choose $\eta$, as described
above. These choices remain fixed
during transfer, while the ID calibration statistics are recomputed for each
target. In the static bridge, S1 already uses all classes; S2 replaces the
bandwidths by their per-bank median and uses uniform anchor weights, S3
normalizes each anchor mean to obtain a radial centroid, and S4 applies the
equivalent maximum-cosine score. Thus scale sharing and candidate expansion are
explicit modeling changes, not consequences of setting $T=0$.

\subsection{Reproducibility notes}

The experiments were executed with Python 3.10.8, PyTorch 2.1.2, torchvision
0.16.2, CUDA 12.1, and an NVIDIA RTX 4090 D. The archived evaluation code
records OpenOOD commit
\nolinkurl{3c35632ee91b54b09d1f085d04f94744cece7d0b}; the CTM completion
metadata records official reference commit
\nolinkurl{3587259bd6a69abd6b4103cb7311ffaa0857d60f}.
Dataset preparation validates
archive integrity, image-list counts, and missing paths. The ImageNet-1K
ResNet-50 file has SHA256
\nolinkurl{0676ba61b6795bbe1773cffd859882e5e297624d384b6993f7c9e683e722fb8a};
the pinned DenseNet-121 file has SHA256
\nolinkurl{42bd2c384fae0b346f977fe9feed1bcfdbd6ea4727478c6c76177d725b218e89}.
The extended audit covers 50 full-dataset ASH/DICE/SHE/RMDS/RankFeat runs.
RankFeat uses full exact SVD, and RMDS streaming statistics match direct tensor
computation within $2.84\times10^{-14}$ in
float64. The CADRef/LogitGap audit adds 22 full-dataset method/model
runs: 11 runs per method across the five dataset/backbone targets. The two
fixed ImageNet-1K backbones are not counted as independent training seeds.
CADRef processes all ID training images for setup, while fixed LogitGap is
training-free; neither uses OOD data for setup or parameter selection.
The accompanying CADRef/LogitGap export includes raw Near-/Far-OOD rows,
mean and standard-deviation summaries, and completion metadata. The expanded
rank table is regenerated from the displayed two-decimal baseline means so
that its tie convention is reproducible from the paper itself.
The score-aligned mechanism study stores its ImageNet-1K selection gate,
locked cross-benchmark transfer gate, pair-decomposition rows, and all
stepwise winner traces. The transfer gate has SHA256
\nolinkurl{1199cf9e330f4e2f623ce7409193306bfc38672daa5bfb1d6c2471624563ec2b};
the winner-switch explanation gate has SHA256
\nolinkurl{f75a83c4a94edd36713ca5e505a01f8a4df28fafd6ae0bf210a8327bd747d9a2}.
Both studies are validation-only and record that no Near-/Far-OOD test loader
was accessed.
The boundary-conditioned stability-probe transfer uses the previously saved
validation caches and formal detectors only. Its locked gate has SHA256
\nolinkurl{c824da104d547c80416682c39e12b55041af190b0312534bd19787f35b06a6bb}
and rejects benchmark-test access. The accompanying result archive has SHA256
\nolinkurl{a2ed6b0dfed5310b343dec71dc7dd5a3d7538d45a3e4100cd77b2d954db34401}.
The complete result matrix, per-run configurations, completion
markers, and summary CSV files are retained in the accompanying reproducibility
archive. The main text and supplementary tables report the formal full-dataset
aggregate experiment families; smoke tests, exploratory score-rule candidates,
and duplicate per-seed raw rows remain available in the audit archive.

\clearpage
\twocolumn
\footnotesize
\setlength{\bibsep}{0pt plus 0.3pt}

\bibliographystyle{apalike} 
\bibliography{bib/Support-vector_networks,bib/Spatial_Pyramid_Matching,bib/Distinctive_image_features_from_scale-invariant_keypoints,bib/Gradient-based_learning_applied_to_document_recognition,bib/imagenet-classification-with-deep-convolutional-neural-networks,bib/Very_deep_convolutional_networks,bib/Going_Deeper_With_Convolutions,bib/Deep_Residual_Learning_for_Image_Recognition,bib/A_ConvNet_for_the_2020s,bib/Algorithms_for_Manifold_Learning,bib/Testing_the_manifold_hypothesis,bib/On_the_local_behavior_of_spaces_of_natural_images,bib/MMCM,bib/Intriguing_properties_of_neural_networks,bib/Explaining_and_Harnessing_Adversarial_Examples,bib/Theory_of_dimensions_finite_and_infinite,bib/The_Intrinsic_Dimension_of_Images_and_Its_Impact_on_Learning,bib/Harnessing_Nonlinearity,bib/Neural_Tangent_Kernel,bib/What_is_the_best_multi-stage_architecture_for_object_recognition,bib/What's_Hidden_in_a_Randomly_Weighted_Neural_Network,bib/The_effective_rank,bib/Regularization_of_Neural_Networks_using_DropConnect,bib/Manifold_Parzen_Windows,bib/EfficientNet,bib/GPipe,bib/Unsupervised_Learning_of_Visual_Representations_by_Solving_Jigsaw_Puzzles,bib/MSP,bib/Principled_Detection_of_Out-of-Distribution_Examples_in_Neural_Networks,bib/Energy-based_Out-of-distribution_Detection,bib/ViM,bib/Out-of-Distribution_Detection_with_Deep_Nearest_Neighbors,bib/Leveraging_Perturbation_Robustness_to_Enhance_Out-of-Distribution_Detection,bib/Extremely_Simple_Activation_Shaping_for_Out-of-Distribution_Detection,bib/Scale,bib/MDS,bib/OpenOOD,bib/GEN,bib/On_Calibration_of_Modern_Neural_Networks,bib/RMDS,bib/Learning_multiple_layers_of_features_from_tiny_images,bib/ReAct,bib/DICE,bib/SHE,bib/RankFeat,bib/CTM,bib/Recent_Geometric_OOD,bib/MaxLogit,bib/LogitGap,bib/Gradient_Short_Circuit,bib/CADRef,bib/GROOD,bib/SPROD,bib/Uncertainty_Aware_Prototype_Refinement,bib/FEVER_OOD}

\begin{thebibliography}{}

\bibitem[Arnold et~al., 1989]{arnold1989mathematical}
Arnold, V.~I., Vogtmann, K., and Weinstein, A. (1989).
\newblock {\em Mathematical methods of classical mechanics}, volume~60.
\newblock Springer.

\bibitem[Cayton, 2008]{cayton2008algorithms}
Cayton, L. (2008).
\newblock Algorithms for manifold learning.
\newblock Technical report, Department of Computer Science \& Engineering, UC
  San Diego.
\newblock \url{https://escholarship.org/uc/item/8969r8tc}.

\bibitem[Djurisic et~al., 2023]{djurisic2023extremely}
Djurisic, A., Bozanic, N., Ashok, A., and Liu, R. (2023).
\newblock Extremely simple activation shaping for out-of-distribution
  detection.
\newblock In {\em The Eleventh International Conference on Learning
  Representations}.
\newblock \url{https://openreview.net/forum?id=ndYXTEL6cZz}.

\bibitem[ElAraby et~al., 2025]{ElAraby2025GROOD}
ElAraby, M., Sahoo, S., Pequignot, Y., Novello, P., and Paull, L. (2025).
\newblock {GROOD}: Gradient-aware out-of-distribution detection.
\newblock {\em Transactions on Machine Learning Research}.

\bibitem[Fefferman et~al., 2016]{fefferman2016testing}
Fefferman, C., Mitter, S., and Narayanan, H. (2016).
\newblock Testing the manifold hypothesis.
\newblock {\em Journal of the American Mathematical Society}, 29(4):983--1049.
\newblock \url{https://doi.org/10.1090/jams/852}.

\bibitem[Gu et~al., 2025]{Gu_2025_ICCV}
Gu, J., Qiao, Z., and Li, Z. (2025).
\newblock Gradient short-circuit: Efficient out-of-distribution detection via
  feature intervention.
\newblock In {\em Proceedings of the IEEE/CVF International Conference on
  Computer Vision}, pages 457--466.

\bibitem[Hendrycks et~al., 2022]{hendrycks2022scaling}
Hendrycks, D., Basart, S., Mazeika, M., Zou, A., Kwon, J., Mostajabi, M.,
  Steinhardt, J., and Song, D. (2022).
\newblock Scaling out-of-distribution detection for real-world settings.
\newblock In {\em Proceedings of the 39th International Conference on Machine
  Learning}, volume 162 of {\em Proceedings of Machine Learning Research},
  pages 8759--8773. PMLR.
\newblock \url{https://proceedings.mlr.press/v162/hendrycks22a.html}.

\bibitem[Hendrycks and Gimpel, 2017]{hendrycks2016baseline}
Hendrycks, D. and Gimpel, K. (2017).
\newblock A baseline for detecting misclassified and out-of-distribution
  examples in neural networks.
\newblock In {\em International Conference on Learning Representations}.
\newblock \url{https://openreview.net/forum?id=Hkg4TI9xl}.

\bibitem[Isaac-Medina et~al., 2025]{IsaacMedina_2025_ICCV}
Isaac-Medina, B. K.~S., Che, M., Gaus, Y. F.~A., Akcay, S., and Breckon, T.~P.
  (2025).
\newblock {FEVER-OOD}: Free energy vulnerability elimination for robust
  out-of-distribution detection.
\newblock In {\em Proceedings of the IEEE/CVF International Conference on
  Computer Vision}, pages 4529--4538.

\bibitem[Ke et~al., 2025]{ke2025pot}
Ke, A., Chen, W., Feng, C., Cao, Y., Xie, X., Zhou, S.~K., and Feng, L. (2025).
\newblock Prototype-based optimal transport for out-of-distribution detection.
\newblock In Kwok, J., editor, {\em Proceedings of the Thirty-Fourth
  International Joint Conference on Artificial Intelligence, {IJCAI-25}}, pages
  3018--3026. International Joint Conferences on Artificial Intelligence
  Organization.
\newblock Main Track.

\bibitem[Lee et~al., 2018]{NEURIPS2018_abdeb6f5}
Lee, K., Lee, K., Lee, H., and Shin, J. (2018).
\newblock A simple unified framework for detecting out-of-distribution samples
  and adversarial attacks.
\newblock In Bengio, S., Wallach, H., Larochelle, H., Grauman, K.,
  Cesa-Bianchi, N., and Garnett, R., editors, {\em Advances in Neural
  Information Processing Systems}, volume~31. Curran Associates, Inc.
\newblock
  \url{https://proceedings.neurips.cc/paper_files/paper/2018/file/abdeb6f575ac5c6676b747bca8d09cc2-Paper.pdf}.

\bibitem[Liang et~al., 2025]{liang2025logitgap}
Liang, J., Hou, R., Hu, M., Chang, H., Shan, S., and Chen, X. (2025).
\newblock Revisiting logit distributions for reliable out-of-distribution
  detection.
\newblock In {\em Advances in Neural Information Processing Systems},
  volume~38.

\bibitem[Liang et~al., 2017]{DBLP:journals/corr/LiangLS17}
Liang, S., Li, Y., and Srikant, R. (2017).
\newblock Principled detection of out-of-distribution examples in neural
  networks.
\newblock {\em CoRR}, abs/1706.02690.
\newblock \url{http://arxiv.org/abs/1706.02690}.

\bibitem[Ling et~al., 2025]{Ling_2025_CVPR}
Ling, Z., Chang, Y., Zhao, H., Zhao, X., Chow, K., and Deng, S. (2025).
\newblock {CADRef}: Robust out-of-distribution detection via class-aware
  decoupled relative feature leveraging.
\newblock In {\em Proceedings of the IEEE/CVF Conference on Computer Vision and
  Pattern Recognition}, pages 4968--4977.

\bibitem[Liu et~al., 2020]{NEURIPS2020_f5496252}
Liu, W., Wang, X., Owens, J., and Li, Y. (2020).
\newblock Energy-based out-of-distribution detection.
\newblock In Larochelle, H., Ranzato, M., Hadsell, R., Balcan, M., and Lin, H.,
  editors, {\em Advances in Neural Information Processing Systems}, volume~33,
  pages 21464--21475. Curran Associates, Inc.
\newblock
  \url{https://proceedings.neurips.cc/paper_files/paper/2020/file/f5496252609c43eb8a3d147ab9b9c006-Paper.pdf}.

\bibitem[Liu et~al., 2023]{Liu_2023_CVPR}
Liu, X., Lochman, Y., and Zach, C. (2023).
\newblock Gen: Pushing the limits of softmax-based out-of-distribution
  detection.
\newblock In {\em Proceedings of the IEEE/CVF Conference on Computer Vision and
  Pattern Recognition (CVPR)}, pages 23946--23955.
\newblock
  \url{https://openaccess.thecvf.com/content/CVPR2023/html/Liu_GEN_Pushing_the_Limits_of_Softmax-Based_Out-of-Distribution_Detection_CVPR_2023_paper.html}.

\bibitem[Nguyen et~al., 2026]{nguyen2026fisherrao}
Nguyen, A., Bertrand, A., Le~H{\'e}garat-Mascle, S., Aldea, E., El-Korso,
  M.~N., Lustrat, R., and Florin, F. (2026).
\newblock Fisher--rao sensitivity for out-of-distribution detection in deep
  neural networks.
\newblock In {\em The Fourteenth International Conference on Learning
  Representations}.

\bibitem[Nguyen et~al., 2023]{nguyen2023ctm}
Nguyen, N.-H., Nguyen, H.-Q., Ta, T.-A., Nguyen-Tang, T., Doan, K.~D., and
  Hoang, T.-T. (2023).
\newblock A cosine similarity-based method for out-of-distribution detection.
\newblock In {\em ICML 2023 Workshop on Spurious Correlations, Invariance, and
  Stability}.
\newblock \url{https://openreview.net/forum?id=QuXApqb8Yl}.

\bibitem[Park et~al., 2023]{park2023nnguide}
Park, J., Jung, Y.~G., and Teoh, A. B.~J. (2023).
\newblock Nearest neighbor guidance for out-of-distribution detection.
\newblock In {\em Proceedings of the IEEE/CVF International Conference on
  Computer Vision}, pages 1686--1695.

\bibitem[Ren et~al., 2021]{DBLP:journals/corr/abs-2106-09022}
Ren, J., Fort, S., Liu, J.~Z., Roy, A.~G., Padhy, S., and Lakshminarayanan, B.
  (2021).
\newblock A simple fix to mahalanobis distance for improving near-ood
  detection.
\newblock {\em CoRR}, abs/2106.09022.
\newblock \url{https://arxiv.org/abs/2106.09022}.

\bibitem[Seleznova et~al., 2026]{seleznova2026gradpca}
Seleznova, M., Chou, H.-H., {Mayrink Verdun}, C., and Kutyniok, G. (2026).
\newblock {GradPCA}: Leveraging {NTK} alignment for reliable
  out-of-distribution detection.
\newblock In {\em The Fourteenth International Conference on Learning
  Representations}.

\bibitem[Song et~al., 2022]{song2022rankfeat}
Song, Y., Sebe, N., and Wang, W. (2022).
\newblock {RankFeat}: Rank-1 feature removal for out-of-distribution detection.
\newblock In {\em Advances in Neural Information Processing Systems},
  volume~35, pages 17885--17898.
\newblock
  \url{https://proceedings.neurips.cc/paper_files/paper/2022/hash/71c9eb0913e6c7fda3afd69c914b1a0c-Abstract-Conference.html}.

\bibitem[Sun et~al., 2021]{sun2021react}
Sun, Y., Guo, C., and Li, Y. (2021).
\newblock React: Out-of-distribution detection with rectified activations.
\newblock In {\em Advances in Neural Information Processing Systems},
  volume~34.
\newblock
  \url{https://proceedings.neurips.cc/paper/2021/hash/01894d6f048493d2cacde3c579c315a3-Abstract.html}.

\bibitem[Sun and Li, 2022]{sun2022dice}
Sun, Y. and Li, Y. (2022).
\newblock {DICE}: Leveraging sparsification for out-of-distribution detection.
\newblock In {\em Computer Vision -- ECCV 2022}, pages 691--708.
\newblock
  \url{https://www.ecva.net/papers/eccv_2022/papers_ECCV/papers/136840680.pdf}.

\bibitem[Sun et~al., 2022]{pmlr-v162-sun22d}
Sun, Y., Ming, Y., Zhu, X., and Li, Y. (2022).
\newblock Out-of-distribution detection with deep nearest neighbors.
\newblock In Chaudhuri, K., Jegelka, S., Song, L., Szepesvari, C., Niu, G., and
  Sabato, S., editors, {\em Proceedings of the 39th International Conference on
  Machine Learning}, volume 162 of {\em Proceedings of Machine Learning
  Research}, pages 20827--20840. PMLR.
\newblock \url{https://proceedings.mlr.press/v162/sun22d.html}.

\bibitem[Wang et~al., 2022]{Wang_2022_CVPR}
Wang, H., Li, Z., Feng, L., and Zhang, W. (2022).
\newblock Vim: Out-of-distribution with virtual-logit matching.
\newblock In {\em Proceedings of the IEEE/CVF Conference on Computer Vision and
  Pattern Recognition (CVPR)}, pages 4921--4930.
\newblock
  \url{https://openaccess.thecvf.com/content/CVPR2022/html/Wang_ViM_Out-of-Distribution_With_Virtual-Logit_Matching_CVPR_2022_paper.html}.

\bibitem[Wei and Liu, 2026]{WeiLiu2026UAPR}
Wei, T. and Liu, M. (2026).
\newblock Uncertainty-aware prototype refinement for out-of-distribution
  detection.
\newblock {\em Neurocomputing}, 699:134310.

\bibitem[Xia and Bouganis, 2024]{xia2024sirc}
Xia, G. and Bouganis, C.-S. (2024).
\newblock Augmenting the softmax with additional confidence scores for improved
  selective classification with out-of-distribution data.
\newblock {\em International Journal of Computer Vision}, 132(9):3714--3752.

\bibitem[Xu et~al., 2024]{ICLR2024_46d943bc}
Xu, K., Chen, R., Franchi, G., and Yao, A. (2024).
\newblock Scaling for training time and post-hoc out-of-distribution detection
  enhancement.
\newblock In Kim, B., Yue, Y., Chaudhuri, S., Fragkiadaki, K., Khan, M., and
  Sun, Y., editors, {\em International Conference on Learning Representations},
  volume 2024, pages 16316--16329.
\newblock
  \url{https://proceedings.iclr.cc/paper_files/paper/2024/file/46d943bc6a15a57c923829efc0db7c7a-Paper-Conference.pdf}.

\bibitem[Zhang et~al., 2023a]{zhang2023she}
Zhang, J., Fu, Q., Chen, X., Du, L., Li, Z., Wang, G., Liu, X., Han, S., and
  Zhang, D. (2023a).
\newblock Out-of-distribution detection based on in-distribution data patterns
  memorization with modern hopfield energy.
\newblock In {\em The Eleventh International Conference on Learning
  Representations}.
\newblock \url{https://openreview.net/forum?id=KkazG4lgKL}.

\bibitem[Zhang et~al., 2023b]{zhang2023openood}
Zhang, J., Yang, J., Wang, P., Wang, H., Lin, Y., Zhang, H., Sun, Y., Du, X.,
  Li, Y., Liu, Z., Chen, Y., and Li, H. (2023b).
\newblock Openood v1.5: Enhanced benchmark for out-of-distribution detection.
\newblock {\em arXiv preprint arXiv:2306.09301}.
\newblock \url{https://arxiv.org/abs/2306.09301}.

\bibitem[Zohrabi et~al., 2025]{Zohrabi2025SPROD}
Zohrabi, R., Hasani, H., Baghshah, M., Rohrbach, A., Rohrbach, M., and Rohban,
  M.~H. (2025).
\newblock Spurious-aware prototype refinement for reliable out-of-distribution
  detection.
\newblock In {\em Advances in Neural Information Processing Systems},
  volume~38.

\end{thebibliography}

\end{document}